%% file: paper.tex
\documentclass[]{bytedance_seed}

\usepackage[toc,page,header]{appendix}

\usepackage{minitoc}

\usepackage{amsmath}
\usepackage{amssymb}
\usepackage{mathtools}
\usepackage{amsthm}
\usepackage{multirow}
\usepackage{makecell}
\usepackage{dsfont}
\usepackage{multicol}

\usepackage{algorithm}
\usepackage{algpseudocode}

\usepackage{listings}
\usepackage{courier}
\usepackage{xcolor}
\title{Expert Race: A Flexible Routing Strategy for Scaling Diffusion Transformer with Mixture of Experts}

\author[1,*]{$\ $ Yike Yuan}
\author[1,2,*]{\quad Ziyu Wang}
\author[1]{\quad Zihao Huang}
\author[1]{\quad Defa Zhu}
\author[1]{\quad Xun Zhou}
\author[2, \dagger]{\quad Jingyi Yu}
\author[1, \dagger]{\quad Qiyang Min}

\affiliation[1]{ByteDance Seed}
\affiliation[2]{ShanghaiTech University}

\contribution[*]{Equal Contribution}
\contribution[\dagger]{Corresponding Authors}

\abstract{
Diffusion models have emerged as mainstream framework in visual generation. Building upon this success, the integration of Mixture of Experts (MoE) methods has shown promise in enhancing model scalability and performance. In this paper, we introduce Race-DiT, a novel MoE model for diffusion transformers with a flexible routing strategy, Expert Race.  By allowing tokens and experts to compete together and select the top candidates, the model learns to dynamically assign experts to critical tokens. Additionally, we propose per-layer regularization to address challenges in shallow layer learning, and router similarity loss to prevent mode collapse, ensuring better expert utilization.
Extensive experiments on ImageNet validate the effectiveness of our approach, showcasing significant performance gains while promising scaling properties.
}

\date{March 26, 2025}
\correspondence{\\ Qiyang Min at  \email{minqiyang@bytedance.com}, \\ Jingyi Yu at \email{yujingyi@shanghaitech.edu.cn}}

\checkdata[Acknowledgement]{We would like to thank Fan Yin, Xudong Sun, and Heng Zhang at ByteDance Seed Team for their support on infrastructure to accelerate training.}

\begin{document}
\maketitle

\newpage
\tableofcontents
\newpage

\input{sections/introduction}
\input{sections/relatedwork}
\input{sections/approach}

\input{sections/experiments}

\clearpage

\bibliographystyle{plainnat}
\bibliography{main}

\clearpage

\beginappendix

\input{sections/appendix}

\end{document}

%% file: sections/introduction.tex
\section{Introduction}

\begin{figure}[H]
        \centering
        \includegraphics[width=0.98\linewidth]{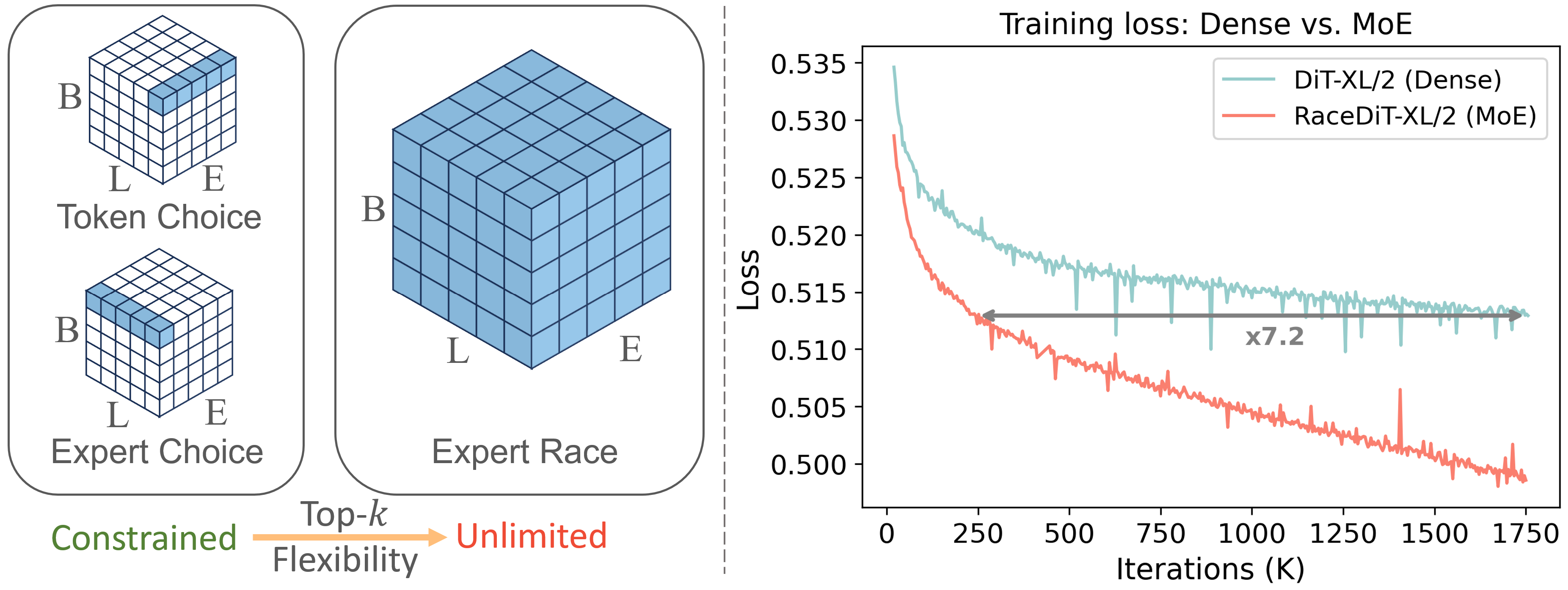}
    \caption{(Left) Our proposed {\bf Expert Race} routing employs Top-$k$ selection over full token-expert affinity logits, achieving the {\bf highest flexibility} compared to prior methods like Token Choice and Expert Choice. \\(Right) Training curve comparisons between DiT-XL~\citep{dit} and ours. Our model, with {\bf equal number of activated parameters}, achieves a {\bf $\mathbf{7.2\times}$ speedup} in iterations when reaching the same training loss.}
    \label{fig:teaser}
\end{figure}

Recent years have seen diffusion models earning considerable recognition within the realm of visual generation. They have exhibited outstanding performance in multiple facets such as image generation~\citep{dalle2,glide,Imagen,sdxl,sd3}, video generation~\citep{W.A.L.T, sora}, and 3D generation~\citep{zhang2024clay, meta_3d_gen}. Thus, diffusion models have solidified their position as a pivotal milestone in the field of visual generation studies. Mimicking the triumph of transformer-based large language models (LLMs), diffusion models have effectively transitioned from U-Net to DiT and its variants. 
This transition yielded not only comparable scaling properties but also an equally successful pursuit of larger models.

In the quest for larger models, the Mixture of Experts (MoE) approach, proven effective in scaling large language models (LLMs)~\citep{mixtral}, exhibits promising potential when incorporated into diffusion models. Essentially, MoE utilizes a routing module to assign tokens among experts (typically, a set of Feed-Forward Networks (FFN)) based on respective scores. This router module, pivotal to MoE's functionality, employs common strategies such as token-choice and expert-choice.

Meanwhile, we observe that the visual signals processed by diffusion models exhibit two distinct characteristics compared to those in LLMs. First, \textbf{visual information tends to have high spatial redundancy.} For instance, significant disparity in information density exists between the background and foreground regions, with the latter typically containing more critical details. Second, \textbf{denoising task complexity exhibits temporal variation across different timesteps.} Predicting noise at the beginning of the denoising process is substantially simpler than predicting noise towards the end, as later stages require finer detail reconstruction. These unique characteristics necessitate specialized routing strategies for visual diffusion models.

Consider these characteristics under the MoE, the presence of a routing module can adaptively allocate computational resources. By \textbf{assigning more experts to challenging tokens} and fewer to simpler ones, we can enhance model utilization efficiency. Previous strategies like expert-choice anticipated this, but their routing design limit the assignment flexibility to image spatial regions without considering temporal denoising timestep complexity.

In this paper, we introduce Race-DiT, a novel family of MoE models equipped with enhanced routing strategies, Expert Race. We find that simply increasing strategy flexibilities greatly boost the model's performance. Specifically, we conduct a ``race'' among tokens from different samples, timesteps, and experts, and select the top-k tokens from all. This method effectively filters redundant tokens and optimizes computational resource deployment by the MoE.

Expert Race introduces a high degree of flexibility in token allocation within the MoE framework. However, there are several challenges when extending DiT to larger parameter scales using MoE. First, we observe that routing in the shallow layers of MoE struggles to learn the assignment, especially with high-noise inputs. We believe this is due to the weakening of the shallow components in the identity branch of the DiT framework. To address this, we propose an auxiliary loss function with layer-wise regularization to aid in learning. Second, considering the substantial expansion of the candidate space, to prevent the collapse of the allocation strategy, we extend the commonly used balance loss from single experts to combinations of experts. This extension is complemented by our router similarity loss, which ensures effective expert utilization by regulating pairwise expert selection patterns. 

To validate the proposed method, we conducted experiments on ImageNet~\citep{deng2009imagenet}, performing detailed ablations on the proposed modules and investigating the scaling behaviors of multiple factors. Results show that our approach achieves significant improvements across multiple metrics compared to baseline methods.

In summary, our main contributions include
\begin{itemize}
    \item Expert Race, a novel MoE routing strategy for diffusion transformers that supports high routing allocation flexibility in both spatial image regions and temporal denoising steps.
    \item Router similarity loss, a new objective that optimizes expert collaboration through router logits similarity, effectively maintaining workload equilibrium and diversifying expert combinations without compromising generation fidelity.
    \item Per-layer Regularization that ensures effective learning in the shallow layers of MoE models. 
    \item Detailed MoE scaling analysis in terms of hidden split and expert expansion provides insights for extending this MoE model to diverse diffusion tasks.
\end{itemize}

\begin{figure}[t!]
\centering
    \includegraphics[width=0.7\textwidth]{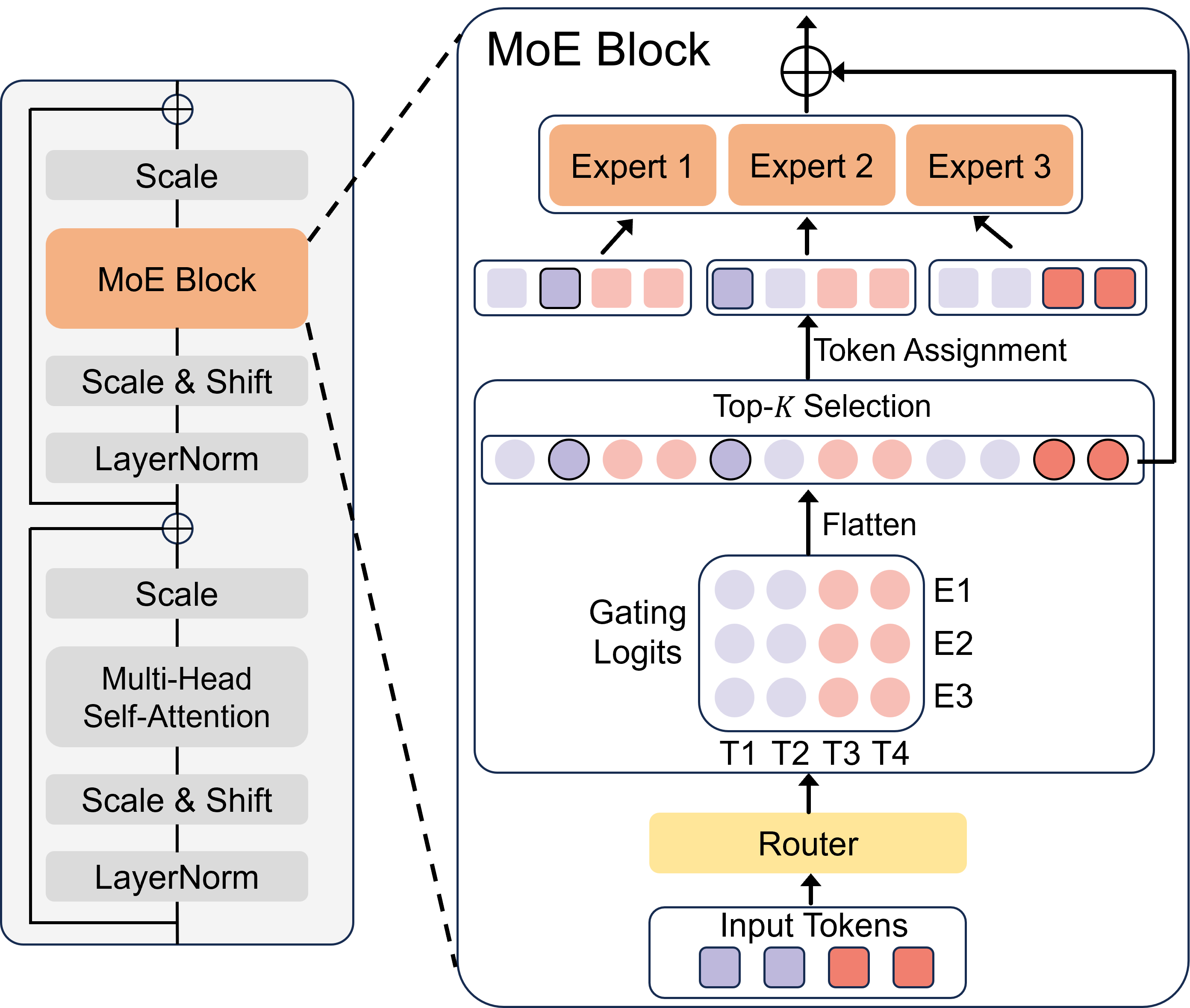}
    \caption{The Race-DiT Architecture. We replace the Multi-Layer Perceptron (MLP) with the MoE block, which consists of a Router and multiple Experts. In Race-DiT, the token assignment is done once for all. Each token can be assigned to any number of experts, and each expert can process any number of tokens (including zero).}
\end{figure}

%% file: sections/relatedwork.tex
\newpage
\section{Related Work}

\subsection{Mixture of Experts}
Mixture of Experts (MoE) improves computational efficiency by activating only a subset of parameters at a time and forcing the other neurons to be zero. Typically, MoE is used to significantly scale up models beyond their current size leveraging the natural sparsity of activations. This technique has been widely applied in LLMs first~\citep{gshard,switch_transformer} and then extended to the vision domain~\citep{vision-moe}. 
The most commonly used routing strategy in MoE is token-choice, in which each token selects a subset of experts according to router scores. For its variants, THOR~\citep{thor} employs a random strategy, BASELayer~\citep{baselayer} addresses the linear assignment problem, HASHLayer~\citep{hashlayer} uses a hashing function, and MoNE~\citep{mone} uses greedy top-k. All of these methods allocate fixed number of experts to each token. DYNMoE~\citep{dynmoe} and ReMoE~\citep{remoe} activates different number of experts for each token by replacing TopK with threshold and using additional regularization terms to control the total budget. 
Also, some auxiliary regularization terms are applied to constrain the model to activate experts uniformly~\citep{zoph2022st, deepseekmoe,aux_loss_free}.
Expert choice~\citep{expert_choice} has been proposed to avoid load imbalance without additional regularizations and enhance routing dynamics, but due to conflicts with mainstream causal attention, it is less commonly applied in large language models (LLMs). 

\subsection{Multiple Experts in Diffusion}
Diffusion follows a multi-task learning framework that share the same model across different timesteps. Consequently, many studies have explored whether performance can be enhanced by disentangling tasks according to timesteps inside the model. Ernie~\citep{ernie} and e-diff~\citep{ediff} manually separate the denoising process into multiple stages and train different models to handle each stage. MEME~\citep{meme} uses heterogeneous models and DTR~\citep{dtr} heuristically partitions along the channel dimension. DyDiT~\citep{dynamic_dit} introduce nested MLPs and channel masks to fit varing complexities across time and spatial dimensions. 
DiT-MoE~\citep{dit_moe}, EC-DiT~\citep{ec_dit}, and Raphael~\citep{raphael} have applied MoE architectures, learning to assign experts to tokens in an end-to-end manner. Compared with previous works, the methods proposed in this paper builds on the MoE but further enhancing its flexibility on dynamically allocate experts on all dimensions to unleash its potential.

%% file: sections/approach.tex
\section{Preliminaries}
Before introducing our MoE design, we briefly review some preliminaries of diffusion models and mixture of experts.
\subsection{Diffusion Models}

Diffusion models~\citep{DDPM} are a class of generative models that sample from a noise distribution and learn a gradual denoising process to generate clean data. It can be seen as an interpolation process between the data sample $x_0$ and the noise $\epsilon$. A typical Gaussian diffusion models formulates the forward diffusion process as
\begin{equation}
x_t = \sqrt{\bar\alpha_t} x_0 + \sqrt{1-\bar\alpha_t}\epsilon
\end{equation}
where $\epsilon\sim\mathcal{N}(0,\mathbf{I})$ is the Gaussian noise and $\bar\alpha_t$ is a monotonically decreasing function from 1 to 0.
Diffusion models use the neural networks to estimate the reverse denoising process
$p_\theta(x_{t-1}|x_t)=\mathcal{N}\left(\mu_\theta(x_t), \Sigma_\theta(x_t)\right)$. They are trained by minimizing the following objectives:
\begin{equation}
    \min_\theta\mathbb{E}_{x_0,t,\epsilon}\left[ \| \mathbf{y} - F_\theta(x_t;c,t) \|^2 \right],
\end{equation}
where $t$ is the timestep which uniformly distributed between $0$ to $T$ , $c$ is the condition information. e.g. class labels, image or text-prompt. The training target $\mathbf{y}$ can be a Gaussian noise $\epsilon$, the original data sample $x_0$ or velocity $v=\sqrt{1-\bar\alpha}\epsilon-\sqrt{\bar\alpha}x_0$.

Early diffusion models used U-net~\citep{cg,LDM} as their backbone. Recently, Transformer-based diffusion models~\citep{dit} with adaptive layer normalization (AdaLN)~\citep{adaln} have become mainstream, showing significant advantages in scaling up.

\subsection{Mixture of Experts}
Mixture-of-Experts (MoE) is a neural network layer comprising a router $\mathcal{R}$ and a set $\{{E_i}\}$ of 
$N_E$ experts, each specializing in a subset of the input space and implemented as FFN. 
The router maps the input $X\in\mathbb{R}^{B\times L\times D}$ into token-expert affinity scores $\mathbf{S}\in\mathbb{R}^{B\times L\times E}$, trailed by a gating function $\mathcal{G}$:
\begin{equation}
    \mathbf{S} = \mathcal{G}(\mathcal{R}(x)).
\end{equation}

The input will be assigned to a subset of experts with top-k highest scores for computation and its output is the weighted sum of these experts' output.
A unified expression is as follows:

\begin{equation}
\vspace{-1em}
\label{eq:gate}
\mathbf{G} =
\begin{cases}
      \mathbf{S}, & \text{if } \mathbf{S} \in\texttt{TopK}\left(\mathbf{S}, \mathcal{K}  \right) \\
      0, &\text{Otherwise}
\end{cases}
\end{equation}

\begin{equation}
    \text{MoE}(X) = \sum_{i\in N_E} G_i(X) * E_i(X)
\end{equation}

where $\mathbf{G}\in\mathbb{R}^{B\times L \times E}$ is the final gating tensor and $\texttt{TopK}(\cdot, \mathcal{K})$ is an operation that build a set with $\mathcal{K}$ largest value in tensor. 

To maintain a constant number of activated parameters while increasing the top-k expert selection, the MoE model often splits the inner hidden dimension of each expert based on the top-k value, named fine-grained expert segmentation~\citep{deepseekmoe}.
In the subsequent discussions, an "{\bf $\mathbf{x}$-in-$\mathbf{y}$}" MoE means there are $y$ candidate experts, with the top-$x$ experts activated, and the hidden dimension of expert's intermediate layer will be divided by $x$.

\subsection{The Rationality of Using MoE in Diffusion Models}
Diffusion models possess several distinctive characteristics. 
\begin{itemize}
    \item Multi-task in nature, tasks at different timesteps predicting the target are not identical. Prior works like e-diff~\citep{ediff} validate this dissimilarity.
    \item Redundancy of image tokens. The information density varies across different regions, leading to unequal difficulties in generation. 
\end{itemize}

Given these traits, MoE presents a suitable architecture for diffusion models. Its routing module can flexibly allocate and combine tokens and experts based on the predicted difficulties.
We consider the allocation process as a distribution of computational resources. More challenging timesteps and complex image patches should be allocated to more experts. Achieving this requires a routing strategy with sufficient flexibility to distribute resources with broader degrees of freedom. Our method is designed following this principle.

\section{Taming Diffusion Models with Expert Race}
\subsection{General Routing Formulation}
\label{sec:route_formulation}

For computational tractability, we decompose the original score tensor $\mathbf{S}$ into two operational dimensions through permutation and reshaping, obtaining matrix $\mathbf{S'} \in \mathbb{R}^{D_A \times D_B}$, where
\begin{itemize}
    \item $D_B$: Size of the expert candidate pool;
    \item $D_A$: Number of parallel selection operations.
\end{itemize}
This dimensional reorganization enables independent top-$k$ selection within each row while preserving cross-row independence.

Following the sparse gating paradigm in \cite{expert_choice,gshard}, we control the MoE layer sparsity through parameter $k$, which specifies the expected number of activated experts per token. To satisfy system capacity constraints, the effective selection size per candidate pool is defined as:
\begin{equation}
    \mathcal{K} = \frac{k}{N_E} \cdot D_B.
\end{equation}

The routing objective, aligning with the optimization framework in \cite{baselayer}, formalizes as the maximization of aggregated gating scores:
\begin{equation}
\label{eq:routing_objective}
    \max \sum_{i=1}^{D_A} \sum_{j \in \mathcal{T}_i} \mathbf{S'}_{i,j},
\end{equation}
where $\mathcal{T}_i$ denotes the set of indices corresponding to the top-$\mathcal{K}$ values in the $i$-th row of $\mathbf{S'}$.

\begin{figure}[th!]
    \centering
    \begin{minipage}[c]{0.5\textwidth}
        \centering
        \includegraphics[width=\textwidth]{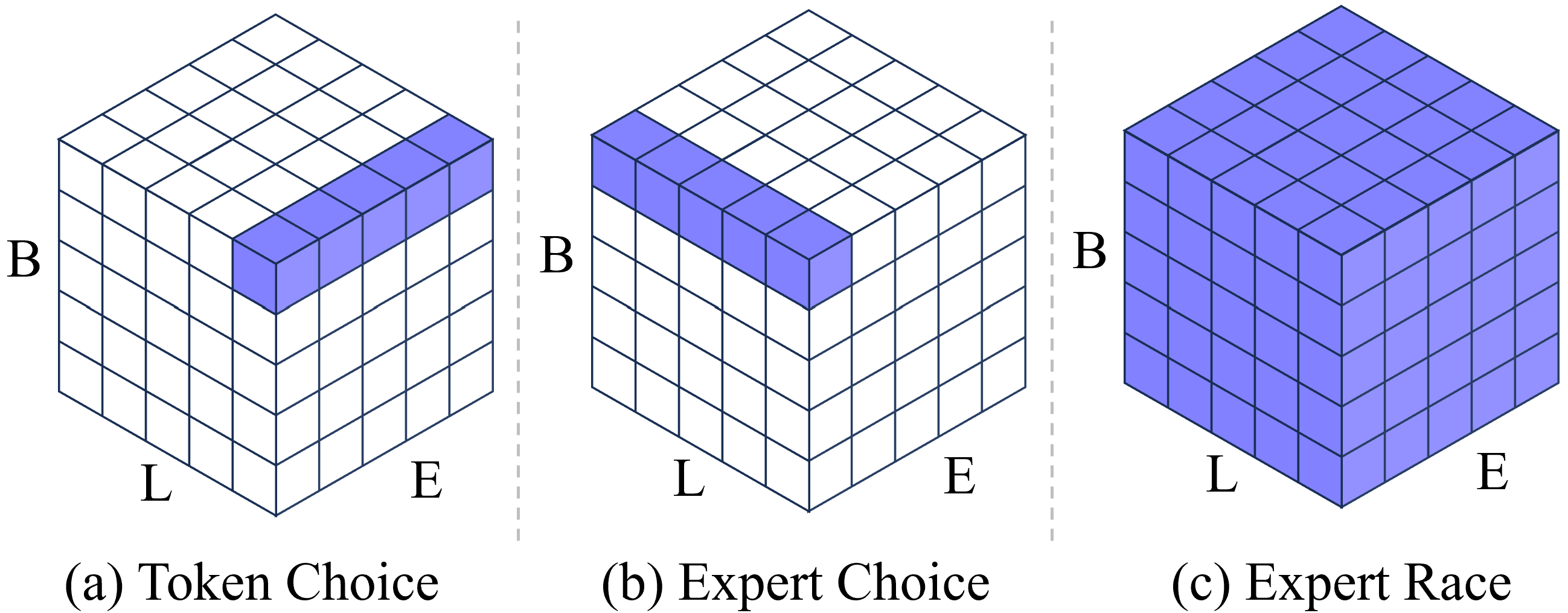}
    \end{minipage}
    \hfill
    \begin{minipage}[c]{0.49\textwidth}
        \centering
        \small
        \begin{tabular}{l|ccc}
        \hline
        \bf{Method} & $D_A$   & $D_B$ & $\mathcal{K}$  \\ 
        \hline \hline
        Token-choice & $B*L$ & $E$ & $k$ \\
        Expert-choice & $B*E$ & $L$ & $k*L/E$ \\
        Expert-Race & $1$ & $B*L*E$ & $B*L*k$ \\\hline\hline
        \end{tabular}
    \end{minipage}
    \caption{Top-$\mathcal{K}$ Selection Flexibility and Specifications.  $B$: batch size; $L$: sequence length; $E$: the number of experts.
        (a) Token Choice selects top-$\mathcal{K}$ experts along the expert dimension for each token. (b) Expert Choice selects top-$\mathcal{K}$ tokens along the sequence dimension for each expert. (c) Expert Race selects top-${\mathcal{K}}$ across the entire set.}
    \label{fig:compare_cube}
\end{figure}

\paragraph{Suboptimal in Conventional Strategies} 
As shown in \cref{fig:compare_cube}, the unified framework generalizes existing routing methods through top-$\mathcal{K}$ selection in $\mathbf{S'}$. However, standard row-wise approaches like Token-Choice and Expert-Choice exhibit inherent sub-optimality. These selection methods struggle to achieve optimal allocation in practice, as the required uniform distribution of top $\mathcal{K} \times D_A$ elements across rows, which is necessary for attaining the theoretical optimum in \cref{eq:routing_objective}, rarely holds with real-world data distributions.

In practical scenarios like image diffusion model training, generation complexity varies across two key dimensions: denoising timesteps ($B$) and spatial image regions ($L$). To address this computational heterogeneity, the routing module must dynamically allocate more experts  to tokens with greater generation demands. However, the token-choice strategy, since $D_A$ is constituted by dimensions $B \& L$, both dimensions will receive an identical amount of activation experts.  
Expert-Choice mitigates this issue but remains constrained by its $L$-dimensional top-$\mathcal{K}$ selection, limiting optimal allocation potential.

\subsection{Expert Race}
To address these limitations, we propose Expert-Race, which performs global top-$\mathcal{K}$ selection across all gating scores in a single routing pass. The "Race" mechanism provides an optimal solution to \cref{eq:routing_objective} by setting $D_A = 1$, ensuring the selected $\mathcal{K}$ elements are globally maximal. This design maximizes router flexibility to learn adaptive allocation patterns, enabling arbitrary expert-to-token assignments and dynamic allocation based on computational demands. However, applying Expert-Race naively presents two challenges.

\noindent \textbf{Gating Function Conflict.} While softmax over the expert dimension is standard for score normalization in existing routing strategies, it disrupts cross-token score ordering in Race. Additionally, applying softmax across the full sequence incurs high computational costs and risks numerical underflow as sequence length grows. We therefore explore alternative activation functions, finding through \cref{tab:routing_and_gating} that the \texttt{identity} $\mathcal{G}(x) = x$ yields improved results.

\noindent \textbf{Training-Inference Mismatch.} Batch-wise candidate aggregation creates a fundamental mismatch between training and inference. During training, samples influence each other's routing selection and timesteps are randomly sampled per batch, whereas inference operates on independent samples with consistent timesteps. Since timesteps directly control noise mixing levels, this inconsistency degrades generation quality and can lead to model failure. At the same time, the mutual influence between samples during routing selection causes unstable inference. To mitigate these effects, we propose a learnable threshold $\tau$ that estimates the $\mathcal{K}$-th largest value through exponential moving average (EMA) updates during training.
\begin{equation}
    \tau \leftarrow m \tau + (1-m) \cdot \frac{1}{D_A} \sum_{i=1}^{D_A} \mathbf{S'}_{i,\mathcal{K}},
\end{equation}
where $\mathbf{S'}_{i,\mathcal{K}}$ represents the $\mathcal{K}$-th largest element in the $i$-th row of $\mathbf{S'}$. This adaptive threshold is directly applied during inference, ensuring sample independence and consistent performance.

\begin{algorithm}[t]
\caption{Pytorch-style Pseudocode of Expert-Race}
\begin{lstlisting}[language=Python]
# m: momentum
# tau: ema updated threshold
# x: input of shape (B, L, D)
# experts: a list of FFN
# router: a linear layer

# Compute router logits for each token
logits = router(x)   # (B, L, E)    
score = logits.flatten()
gates = nn.Identity()(logits) # activation
expect_k = B * L * k

# Get kthvalue and update threshold
if training:
    kth_val =  -(torch.kthvalue(-score, k=expect_k).values)   # Largest K-th
    mask = score >= kth_val
    tau = m * tau + (1. - m) * kth_val
else:
    mask = score >= tau

# Process tokens by each expert
expert_outputs = torch.stack([expert(x) for expert in experts], dim=-1)

# Aggregate the output by mask
gates = gates * mask.reshape(gates.shape)
output = torch.sum(gates.unsqueeze(-2) * expert_outputs, dim=-1)

\end{lstlisting}
\label{alg}
\end{algorithm}

\textbf{Pseudocode.}
We provide core pseudocode in PyTorch style in \cref{alg}, illustrating how the expert selects the k-th largest logits and updates the threshold. Our algorithm is easy to implement, requiring only minor modifications to the existing MoE framework. 

\begin{figure}[t]
    \centering
    \includegraphics[width = \textwidth]{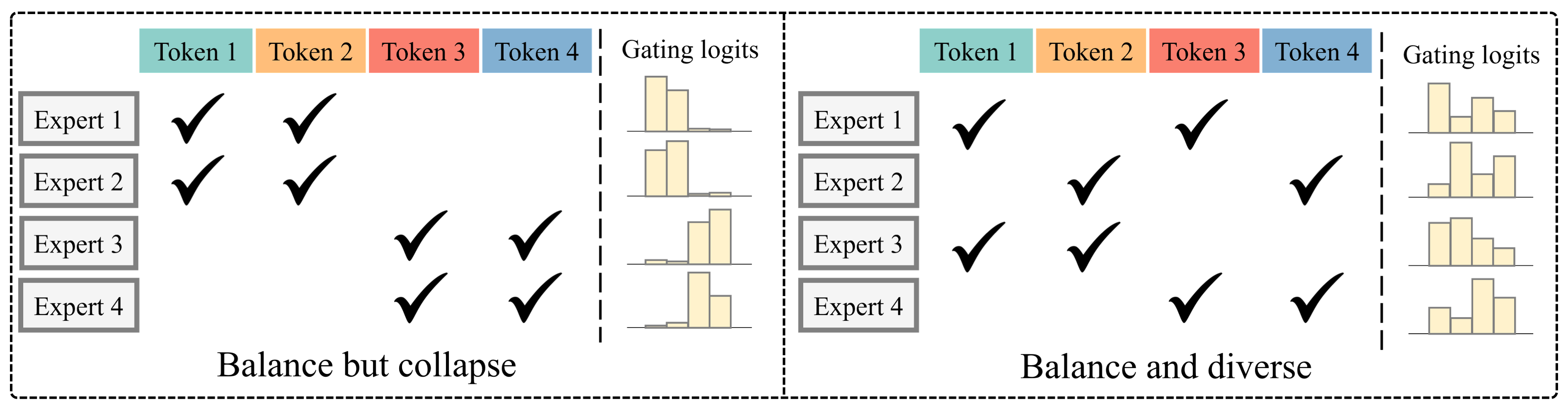}
    \caption{Toy examples of token assignment. Both of the two cases show perfect load balance that each expert process two tokens. But in the case above, experts 1 and 2 are assigned the same token, as are experts 3 and 4, where the 2-in-4 MoE collapse into 1-in-2. The example below shows a more diverse assignment, making full use of the expert specialization.}
    \label{fig:blc_limitation}
\end{figure}

\section{Load Balancing via Router Similarity Loss}
In MoE systems, balanced token allocation across experts remains a critical engineering challenge. For our proposed Race strategy, the increased policy flexibility imposes greater demands on routing balancing.

\subsection{Mode Collapse in Balancing Loss.}  The conventional balancing loss~\citep{shazeer2017moe, switch_transformer}, originally designed for token-choice, promotes load balance by enforcing uniform token distribution across experts, thereby preventing dominance by a small subset of experts. However, by only constraining the marginal distribution of scores per expert, this approach fails to prevent collapse between experts with similar selection rules. As shown in \cref{fig:blc_limitation}, if multiple experts follow the same rules for selecting tokens, they are downgraded to one wider expert. Although such configurations satisfy balance loss constraints, they undermine the specialization benefits of fine-grained expert design~\citep{deepseekmoe}, ultimately degrading overall performance.





\subsection{Router Similarity Loss.}
To tackle this issue, we propose maximizing expert specialization by promoting pairwise diversity among experts. Specifically, inspired by ~\citep{barlowtwins}, we compute cross-correlation matrices and minimize their off-diagonal elements to encourage expert differentiation.  Given the router logits $S \in \mathbb{R}^{(B\times L) \times E}$, we apply softmax along the expert dimension to obtain normalized probabilities $P$, and compute two correlation matrices
\begin{equation}
\label{eq:matrix}
    \begin{aligned}
        M' = M^T M, \quad P' = P^T P
    \end{aligned}
\end{equation}
where $M$ is the indicator matrix that $M_{i,j} = 1$ if expert $j$ selects the $i$-th token and 0 otherwise.

Then, we define the \textbf{router similarity loss}:
\begin{equation}
\small
\label{eq:sim_loss}
\begin{aligned}
\mathcal{L}_{\text{sim}}
&=
\frac{1}{T} \sum_{\substack{i,j \in [1,E]}} W(i,j) \cdot P'_{i,j}
\end{aligned}
\end{equation}

where \( W(i,j) \) is the weighting function defined as:
\begin{equation}
\label{eq:sim_loss_weight}
    W(i,j) = 
    \begin{cases}
        \frac{M'_{i,j}}{\sum_{i=j} M'_{i,j}}\cdot E, & \text{if}\ i = j \\
        \frac{M'_{i,j}}{\sum_{i \neq j} M'_{i,j}}\cdot (E^2-E), & \text{if}\ i \neq j
    \end{cases}
\end{equation}

In more detail, the off-diagonal elements denote the similarity between each pair of experts based on token selection patterns in the current batch. From a probabilistic perspective, $P'_{i,j}$ captures the joint probability of a token being routed to both expert $i$ and $j$. This formulation regularizes consistent co-selection patterns across experts while promoting diverse expert combinations. Regarding the diagonal elements, $P'_{i,i}$ represents a geometric mean version of the balance loss, effectively encouraging individual expert utilization (see \cref{app:sim_loss} for detailed analysis).

\section{Per-Layer Regularization for Efficient Training}

\begin{figure}[h]
    \centering
    \includegraphics[width = 0.65\textwidth]{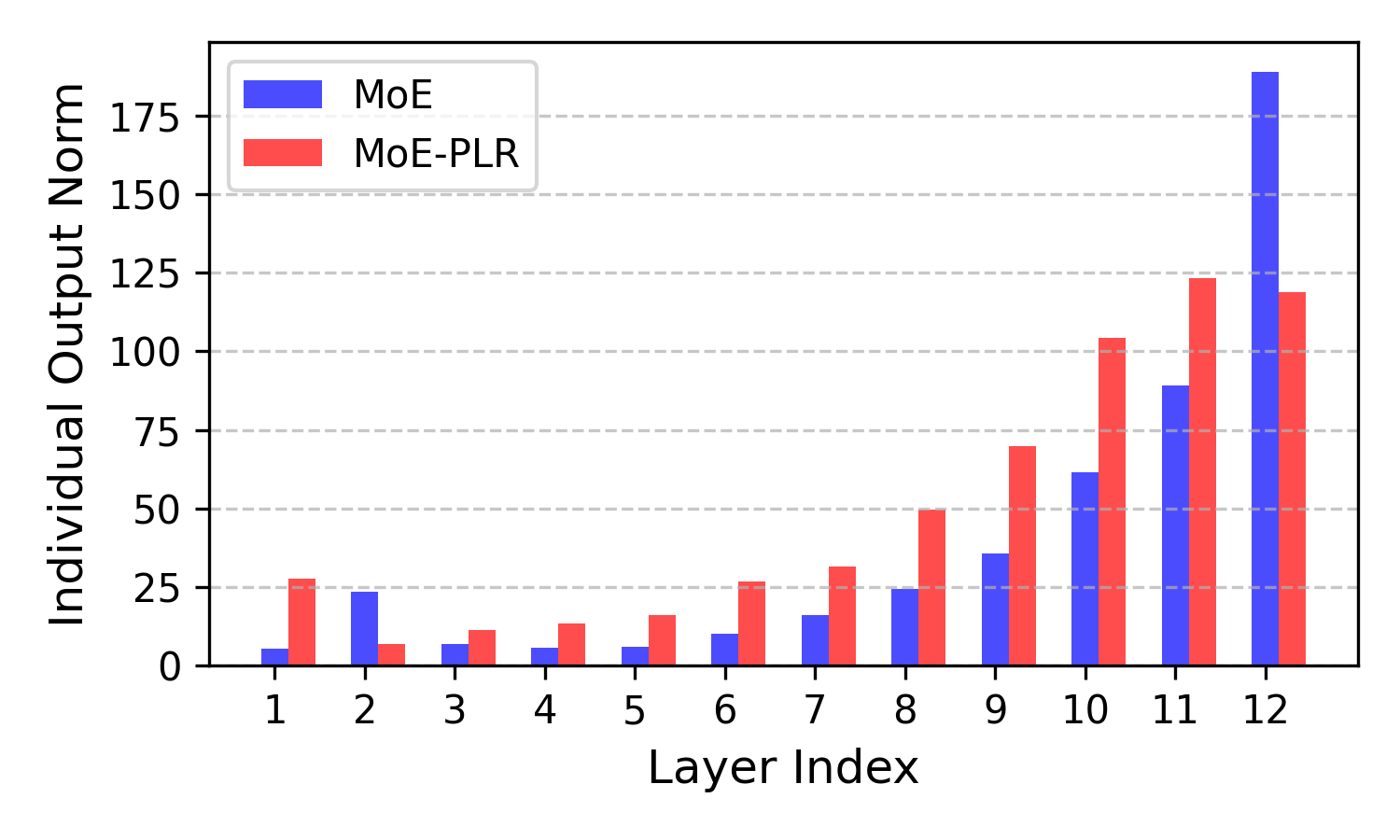}
    \caption{The norm of each block's output before added to the shortcuts. The output norm increases rapidly in deep layers, resulting in the weakening of shallow-layer components. This issue is alleviated with our proposed per-layer regularization.}
    \label{fig:output_norm}
\end{figure}

DiT employs adaptive layer normalization (adaLN) when introducing conditions. In the pre-normalization (pre-norm) architecture, we observe that adaLN progressively amplifies the outputs of deeper layers. This causes the output magnitudes of shallow layers to be relatively diminished, as illustrated in \cref{fig:output_norm}. This imbalance results in the learning speed of shallow layers lagging behind that of deeper layers, which is detrimental to the MoE training process.
This imbalance has both advantages and disadvantages. On one hand, the outputs from deeper layers are more accurate, and their larger magnitudes make them less susceptible to the substantial noise present in shallow layers, facilitating more precise regression results. On the other hand, due to the presence of the normalization module in the final layer, the component of shallow layers in the residuals is diminished, posing a risk of gradient vanishing and resulting in  the learning speed of shallow layers lagging behind that of deeper layers, which is detrimental to the MoE training process.

To mitigate this issue, we introduce a pre-layer regularization that enhances gradients in a supervised manner without altering the core network structure. Specifically, given the hidden output \( \mathbf{h}_l \) from the \( l \)-th layer, we add a projection layer \( \mathcal{H}: \mathbb{R}^{L' \times d} \rightarrow \mathbb{R}^{L \times d} \) to predict the final target \( \mathbf{y} \in \mathbb{R}^{L \times d} \), where \( L \) and \( L' \) represent the number of patches before and after the patchify operation~\citep{}. The pre-layer loss is defined as:
\begin{equation}
\mathcal{L}_{\text{PLR}} = \mathbb{E}_{x_0, t, \epsilon, l}\left[\frac{1}{N} \sum_{i=1}^{N} \left \| \mathbf{y}^{[n]} - \mathcal{H}(\mathbf{h}_l)^{[n]} \right \|^2 \right]
\end{equation}
where \( N \) is the total number of patches, and \( n \) is the patch index. In our implementation, the projection layer is integrated into the MLP router (see \cref{app:reg} for details). By supervising the projection layer's predictions against final targets, we enhance shallow layer contributions during training, improving overall MoE performance.

%% file: sections/experiments.tex
\section{Experiments}

\subsection{Implementation Details}
Following training configurations in~\citep{dit}, we conduct experiments on ImageNet~\citep{deng2009imagenet}, employ AdamW optimizer, set batch size to 256, and use constant learning rate 1e-4 without weight decay for models of any size.
During initialization, we use zero-initialization on all adaLN layers, and xavier initialization of uniform distribution on all linear layers. 
When training MoE, we substitute all FFNs with a MoE block and make sure they activate same amount of parameters.
Specially, we set a smaller initialization range for each expert, enlarging the inner dimension by a factor of k to make the intialization range the same with its dense counterpart. We employ our per-layer regularization with weight 1e-2 and router similarity loss with 1e-4. 
We maintain an exponential moving average (EMA) of model weights over training and report all results using the EMA model.

The metrics used include FID~\citep{fid}, CMMD~\citep{cmmd}, and CLIP Score~\citep{clip}. 
We present a series of MoE size configurations, denoted as 
k-in-E where E represents the total number of experts and k indicates the number of average activated experts. Additionally, we set the inner hidden dimension of each expert to be $1/k$ of its dense counterpart to ensure that the number of activated parameters remains the same.
In all ablation studies, we train DiT-B and its 2-in-8 MoE variant for 500K iterations unless specified otherwise.

\newpage

\begin{figure*}[th!]
    \includegraphics[width=1.0\textwidth]{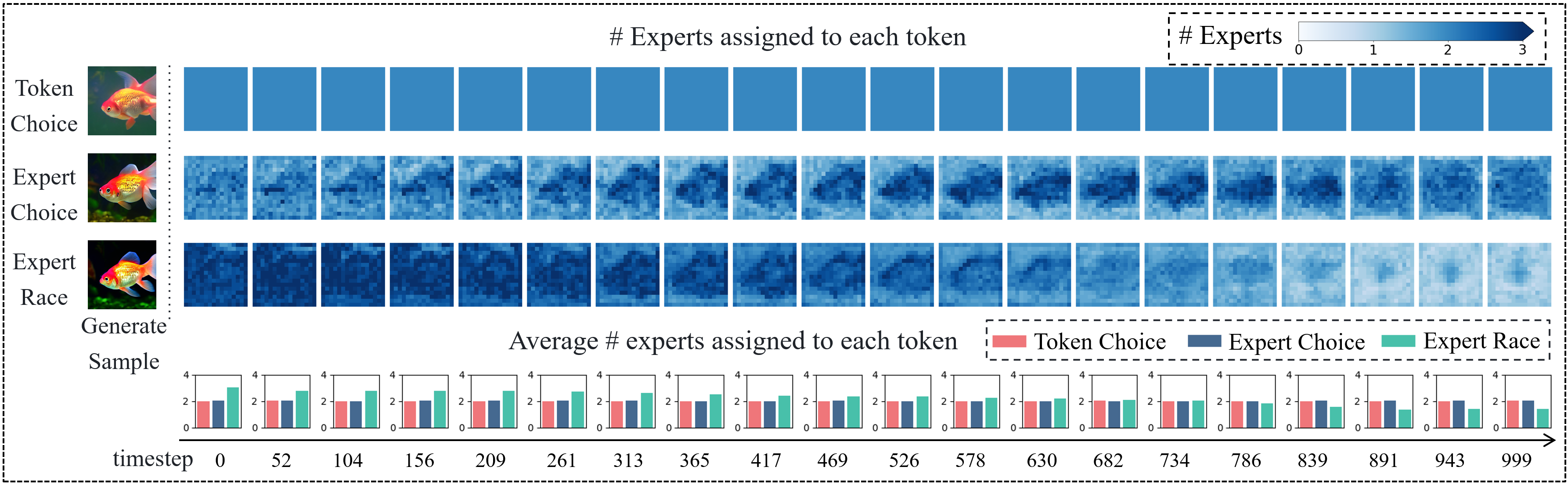}
    \caption{Average token allocation at different time steps. Expert-Race assigns more experts to the more complex denoising time steps, which occur at lower timestep indices that handle finer-grain image details.}
    \label{fig:allocation}
\end{figure*}

\begin{table}[h!]
\small
\centering
\caption{Ablation study on routing strategy and gating function.}
\begin{tabular}{l|c|ccc}
\hline
\multirow{1}{*}{\bf{Routing}} & \multirow{1}{*}{\bf{Gating}}   & \multirow{1}{*}{\bf{FID$\downarrow$}} & \multirow{1}{*}{\bf{CMMD$\downarrow$}} & \multirow{1}{*}{\bf{CLIP$\uparrow$}} \\\hline \hline
Token Choice             & \multirow{3}{*}{softmax}  & 17.28                & .7304                 & 21.87                 \\
Expert Choice            &                           & 16.71                & .7267                 & 21.95                 \\
Expert Race              &                           & 16.47                & .7104                 & 21.97                 \\ \hline
Token Choice             & \multirow{3}{*}{sigmoid}  & 15.25                & .6956                 & 22.09                 \\
Expert Choice            &                           & 15.73                & .6821                 & 22.06                 \\
Expert Race              &                           & 13.85                & .6471                 & 22.23                 \\ \hline
Token Choice             & \multirow{3}{*}{identity} & 15.98                & .6938                 & 22.01                 \\
Expert Choice            &                           & 15.70                & .6963                 & 22.04                 \\
Expert Race              &                           & \textbf{13.66}                & \textbf{.6317}                 & \textbf{22.25}   \\
\hline\hline
\end{tabular}
\label{tab:routing_and_gating}
\end{table}

\subsection{Routing Strategy}
Expert racing enables extensive exploration within the logit space during training, enabling complexity-aware expert allocation across diffusion timesteps. As shown in \cref{fig:allocation}, along the timestep dimension, our method initially uses fewer experts at the beginning of denoising (higher timestep indices) and dynamically assigns more experts to tokens at timesteps requiring higher image detail (lower timestep indices). 
Along the spatial dimension, the assignment of computation follows a "diffusion process" from the center of image to significant object and then to the entire image. This indicates that the model first focuses on object construction and subsequently refines the details.
In contrast, both token-choice and expert-choice strategies maintain fixed average expert allocations per timestep, lacking the temporal dynamic allocation capability. 
From the result proposed in \cref{tab:routing_and_gating}, expert race outperforms other routing strategies by a significant margin.
Within the framework proposed in \cref{sec:route_formulation}, we conducted ablation studies on routing strategies for combinations of different dimensions. See \cref{sec:more_routing} for more results.

\subsection{Gating Function}
\cref{tab:routing_and_gating} shows that identity gating outperforms both softmax and sigmoid variants. In this experiment, we isolate other components (learnable threshold and regularizations) to verify the impact of the gating function on performance. We found that identity gating outperforms softmax and sigmoid under expert-race, and it enhances both token-choice and expert-choice compared to softmax. Under expert-race, both sigmoid and identity significantly outperform softmax. We attribute this to the fact that softmax normalizes scores of each token along the expert dimension, disrupting the partial ordering of scores across different tokens. In contrast, sigmoid and identity preserve this partial ordering, ensuring that important token-expert combinations are selected, thereby improving performance.

\newpage
\subsection{Load Balance}
\cref{tab:balancing} compares our proposed router similarity loss with conventional load balancing loss~\citep{deepseekmoe}.
This setting is similar to the gating function ablation above, but here the MoE is 4-in-32 to further observe the impact of load balancing. The \texttt{MaxVio}~\citep{wang2024auxiliary} metric measures how much the most violated expert exceeds its capacity limit. Combination Usage Ratio, abbreviated as \texttt{Comb}, 
estimates the proportion of activations for each pairwise combination of experts (higher is better).
See \cref{app:combine} for details.

The weights for both the load balance loss and the router similarity loss are set to 1e-4, as this configuration yielded the highest generation quality in our ablation experiments.
Results show our method improves expert combination ratio, image generation quality, and load-balancing performance.

\cref{fig:combination} further demonstrates the evaluation of MoE configurations across multiple scales (4-in-{16,32,64,128}), highlighting our approach's capability to diversify expert activation pattern compared to existing methods.

\begin{minipage}[c]{0.49\textwidth}
    \centering
    \small
    \captionof{table}{Load balance for 4-in-32 MoE.}
    \begin{tabular}{l|ccc}
    \hline
    \bf{Setting}                    & \bf{FID$\downarrow$}    & \bf{MaxVio$\downarrow$} & \bf{Comb$\uparrow$}  \\ \hline \hline
    No Constraint               & 11.38  & 6.383 & 18.98\\
    Balance Loss                & 11.67  & 2.052 & 72.36\\
    Router Similarity       & \bf{10.77}  & \bf{0.850} & \bf{83.10}\\
    \hline\hline
    \end{tabular}
    \label{tab:balancing}
\end{minipage}
\begin{minipage}[c]{0.49\textwidth}
    \centering
    \includegraphics[width=\textwidth]{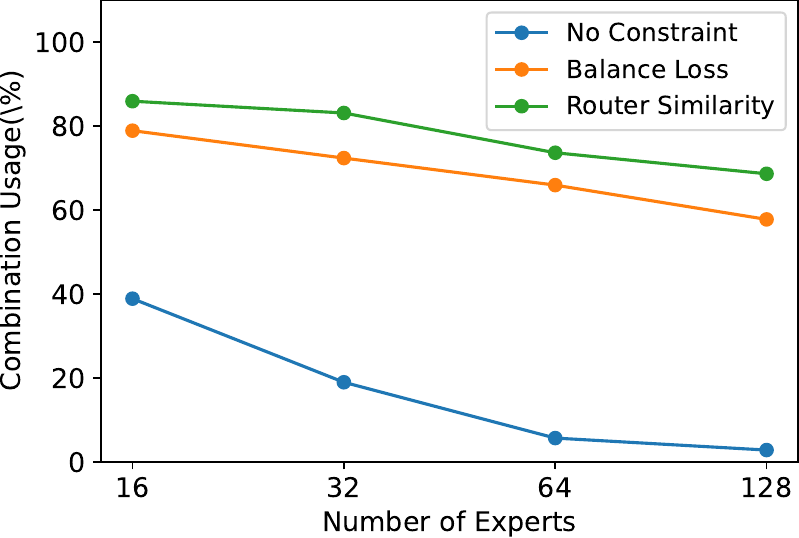}
    \captionof{figure}{Combination usage comparison between different number of experts.}
    \label{fig:combination}
\end{minipage}

\subsection{Core Components}
\cref{tab:component} demonstrates the improvements brought by each component. Starting from the baseline of expert racing with softmax gating, we incrementally added identity gating, learnable thresholds, per-layer regularization, and router similarity loss.
Notably, FID and CMMD consistently decrease and CLIP score increases with the addition of each technique, indicating enhancements in image quality and alignment with conditional distributions.

\begin{table}[h!]
\centering
\small
\caption{Ablation study of core components.}
\begin{tabular}{l|cccc}
\hline
\bf{Setting}                    & \bf{FID$\downarrow$}   & \bf{CMMD$\downarrow$}  & \bf{CLIP$\uparrow$}  \\ \hline \hline
Expert Race (softmax)    & 16.47 & .7104 & 21.97 \\
+ Identity Gating          & 13.66 & .6317 & 22.25 \\
+ Learnable Threshold      & 11.56 & .5863 & 22.56 \\
+ Per-Layer Reg.           & 8.95 & .4847 & 22.94 \\
+ Router Similarity        & \bf{8.03} & \bf{.4587} & \bf{23.09} \\ \hline\hline
\end{tabular}
\label{tab:component}
\end{table}

\newpage
\subsection{Scaling Law}
\begin{figure}[H]
\vspace{-1em}
        \centering
        \includegraphics[width=.9\linewidth]{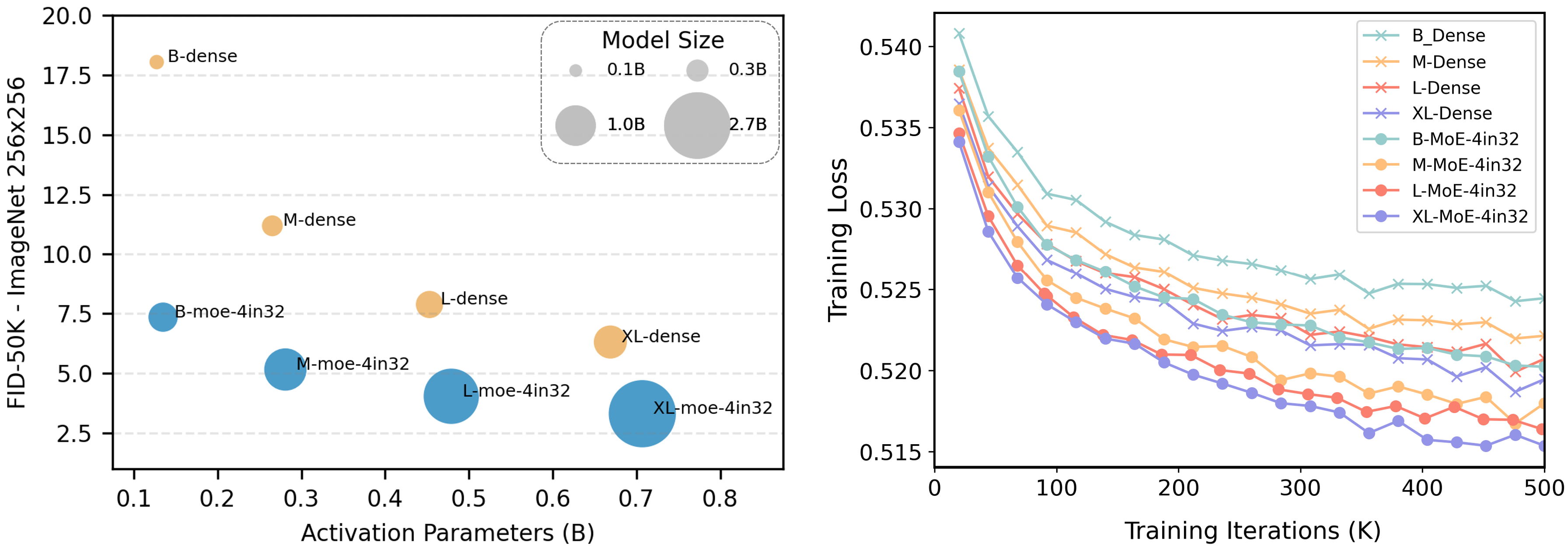}
\vspace{-0.5em}
    \caption{A comparison between Dense and our MoE models. Our MoE models consistently outperform their Dense counterparts across the FID and training curves. Notably, the MoE model with activation size \textbf{\texttt{M}} shows better performance compared to the Dense model scaled to size \textbf{\texttt{XL}}.}
    \label{fig:dense_vs_moe_new}
\vspace{-1em}
\end{figure}

We first scale the base model sizes in the full pipeline, as detailed in \cref{tab:model_spec}. We have four settings: Base (B), Medium (M), Large (L), and Extra-large (XL). Under the same configuration we compared our 4-in-32 MoE models with their dense counterparts, noting that their activation parameter counts are nearly identical. The experimental results and \cref{fig:dense_vs_moe_new} demonstrate that our model significantly outperforms the corresponding Dense models given same activation parameter count. Furthermore, our MoE-4in32 model surpasses the XL-Dense model with less than half the total number of parameters, further showcasing the efficiency of our model design.

\begin{table*}[h!]
\footnotesize
\centering
\caption{Model specifications and evaluation results of the comparison between MoE and Dense models.}
\vspace{-1em}
\begin{tabular}{lcc|ccc|ccc}
\hline
\bf{Model Config.}           & \bf{Total Param.} & \bf{Activate} & \bf{\# Layers} & \bf{Hidden} & \bf{\# Heads} & \bf{FID$\downarrow$}      & \bf{CMMD$\downarrow$}     & \bf{CLIP$\uparrow$}     \\ \hline\hline
B/2-Dense      & 0.127B       & 0.127B          & 12    & 768    & 12   &  18.03       & .7532         & 21.83         \\
M/2-Dense      & 0.265B       & 0.265B          & 16    & 960    & 16   &  11.18       & .5775         & 22.56         \\
L/2-Dense      & 0.453B       & 0.453B          & 24    & 1024   & 16   &  7.88        & .4842         & 23.00         \\
XL/2-Dense     & 0.669B       & 0.669B          & 28    & 1152   & 16   &  6.31        & .4338         & 23.27         \\ \hline
B/2-MoE-4in32  & 0.531B       & 0.135B          & 12    & 768    & 12   &  7.35        & .4460         & 23.15         \\
M/2-MoE-4in32  & 1.106B       & 0.281B          & 16    & 960    & 16   &  5.16        & .3507         & 23.50         \\
L/2-MoE-4in32  & 1.889B       & 0.479B          & 24    & 1024    & 16   &  4.04        & .2775         & 24.12         \\
XL/2-MoE-4in32  & 2.788B       & 0.707B          & 28    & 1152    & 16   &  \bf{3.31}       & \bf{.1784}         & \bf{24.68}         \\
 \hline\hline
\end{tabular}
\label{tab:model_spec}
\end{table*}

\begin{figure}[h!]
    \vspace{-1em}
\centering
    \includegraphics[width=\textwidth]{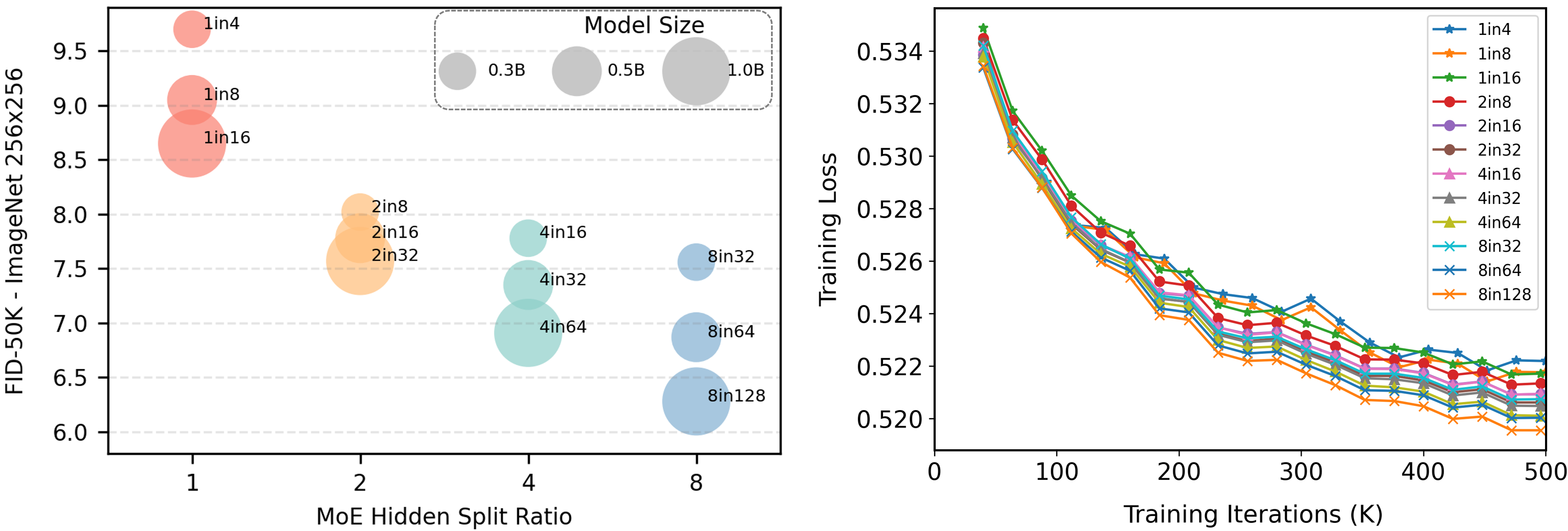}
    \vspace{-2em}
    \caption{Scaling results of DiT-B in different MoE configurations. Our method demonstrates linear performance improvement when varying expert split ratios and increasing the number of candidate experts.}
    \label{fig:scaling}
\end{figure}

We further expand the model's size while \textbf{maintaining the same number of activation parameters} under the model configurations (\textbf{B/2-MoE}). The model expansion is achieved by varying the hidden split ratios of experts and increasing the number of candidate experts. 
As shown in \cref{fig:scaling} and \cref{table:scaling}, increasing both the number of candidate experts and splitting the hidden dimensions of MoE leads to improvement in performance, highlighting the potential of our MoE architecture for scaling up.

\begin{table}[h!]
    \footnotesize
    \centering
    \caption{Evaluation results of different MoE configurations with the same number of activation parameters.}
    \vspace{-1em}
    \begin{tabular}{lcc|ccc}
    \hline
    \bf{Config.} & \bf{Hidden Split} & \bf{Total Param.}   & \bf{FID$\downarrow$}  & \bf{CMMD$\downarrow$} & \bf{CLIP$\uparrow$} \\ \hline  \hline
    1-in-4                           & \multirow{3}{*}{1}               & 0.297B               &     9.70& .5200   &22.82   \\
    1-in-8                           &                                  & 0.524B               &     9.05& .4976   &22.91   \\
    1-in-16                          &                                  & 0.978B               &     8.65& .5019   &22.92   \\ \hline
    2-in-8                           & \multirow{3}{*}{2}               & 0.297B               &     8.03& .4587   &23.09   \\
    2-in-16                          &                                  & 0.524B               &     7.78& .4607   &23.06   \\
    2-in-32                          &                                  & 0.977B               &     7.57& .4483   &23.12   \\ \hline
    4-in-16                          & \multirow{3}{*}{4}               & 0.297B               &     7.78& .4628   &23.09   \\
    4-in-32                          &                                  & 0.524B               &     7.35& .4460   &23.15   \\
    4-in-64                          &                                  & 0.977B               &     6.91& .4244   &23.21   \\ \hline
    8-in-32                          & \multirow{3}{*}{8}               & 0.297B               &     7.56& .4516   &23.11   \\
    8-in-64                          &                                  & 0.524B               &     6.87& .4263   &23.24   \\
    8-in-128                         &                                  & 0.977B               &     \bf{6.28}& \bf{.4015}   & \bf{23.35}   \\ \hline  \hline
    \end{tabular}
    \label{table:scaling}
\end{table}

\newpage

\subsection{Extended Routing Strategy}
\label{sec:more_routing}

\begin{figure*}[h]
    \includegraphics[width=1.0\textwidth]{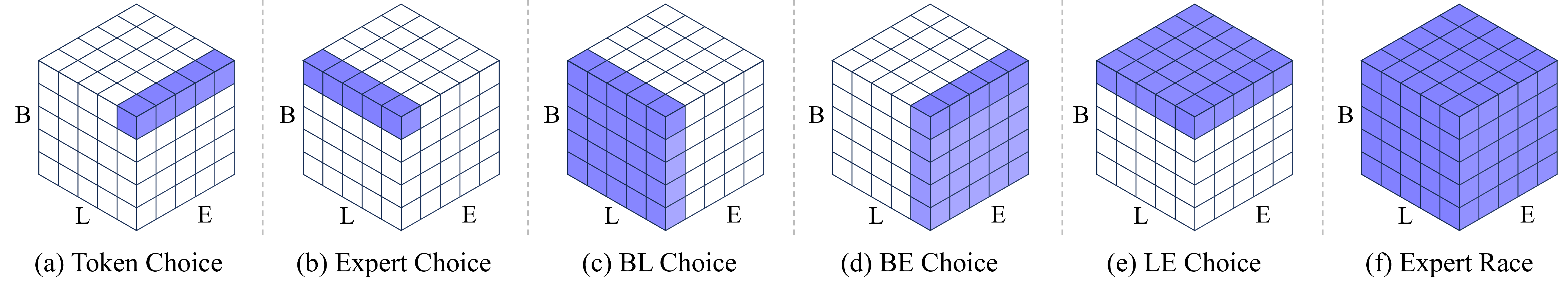}
    \vspace{-2em}
    \caption{Top-$\mathcal{K}$ selection flexibility in more extended routing strategies.}
    \label{fig:allrouting}
\end{figure*}

We extend the token-choice and expert-choice routing strategies by introducing varying degrees of routing selection flexibility, aiming to investigate how training-stage router freedom impacts final model performance. All compared methods are trained for 500K iterations under identical configurations: a full pipeline with DiT-B/2-MoE-2-in-8 architecture. To ensure experimental consistency, all approaches employ learnable thresholds for inference queries. Specifically, we develop three new strategies: BL-Choice, BE-Choice, and LE-Choice, obtained through pairwise combinations of selection dimensions - Batch (B), Sequence Length (L), and Expert count (E), as illustrated in \cref{fig:allrouting}. Experimental results in \cref{tab:bl_be_le} demonstrate that strategies with higher training selection freedom consistently outperform conventional fixed-dimension top-k selection approaches (such as token-choice/expert-choice). Notably, our Expert Race achieves the best performance across all evaluated routing strategies.

\begin{table}[h!]
\centering
    \footnotesize
\caption{Design Choices and Evaluation Results of Different Routing Strategies}
\vspace{-1em}
\begin{tabular}{l|cccccc}
    \hline
     & \bf{Token Choice} & \bf{Expert Choice} & \bf{BL Choice} & \bf{BE Choice} & \bf{LE Choice} & \bf{Expert Race} \\ \hline \hline
$D_A$           & $B*L$   & $B*E$   & $E$         & $L$     & $B$     & $1$ \\
$D_B$           & $E$     & $L$     & $B*L$       & $B*E$   & $L*E$   & $B*L*E$ \\
$\mathcal{K}$   & $k$     & $k*L/E$ & $B*L*k/E$   & $B*k$   & $L*k$   & $B*L*k$ \\ \hline \hline
     
FID$\downarrow$  &      9.50        &  10.13             & 9.08      & 8.28      & 8.89      & \bf{8.03}        \\
CMMD$\downarrow$ &     .5202     &  .5639        & .5145     & .4636    & .4871     & \bf{.4587}      \\
CLIP$\uparrow$ &      22.81        &    22.73           & 22.87     & 23.05     & 22.99       & \bf{23.09}       \\ 
\hline \hline
\end{tabular}
\label{tab:bl_be_le}
\end{table}

\newpage
\subsection{More Results on ImageNet $\mathbf{256\times256}$}

\begin{figure*}[h!]
\centering
    \vspace{-0.5em}
    \includegraphics[width=1.0\textwidth]{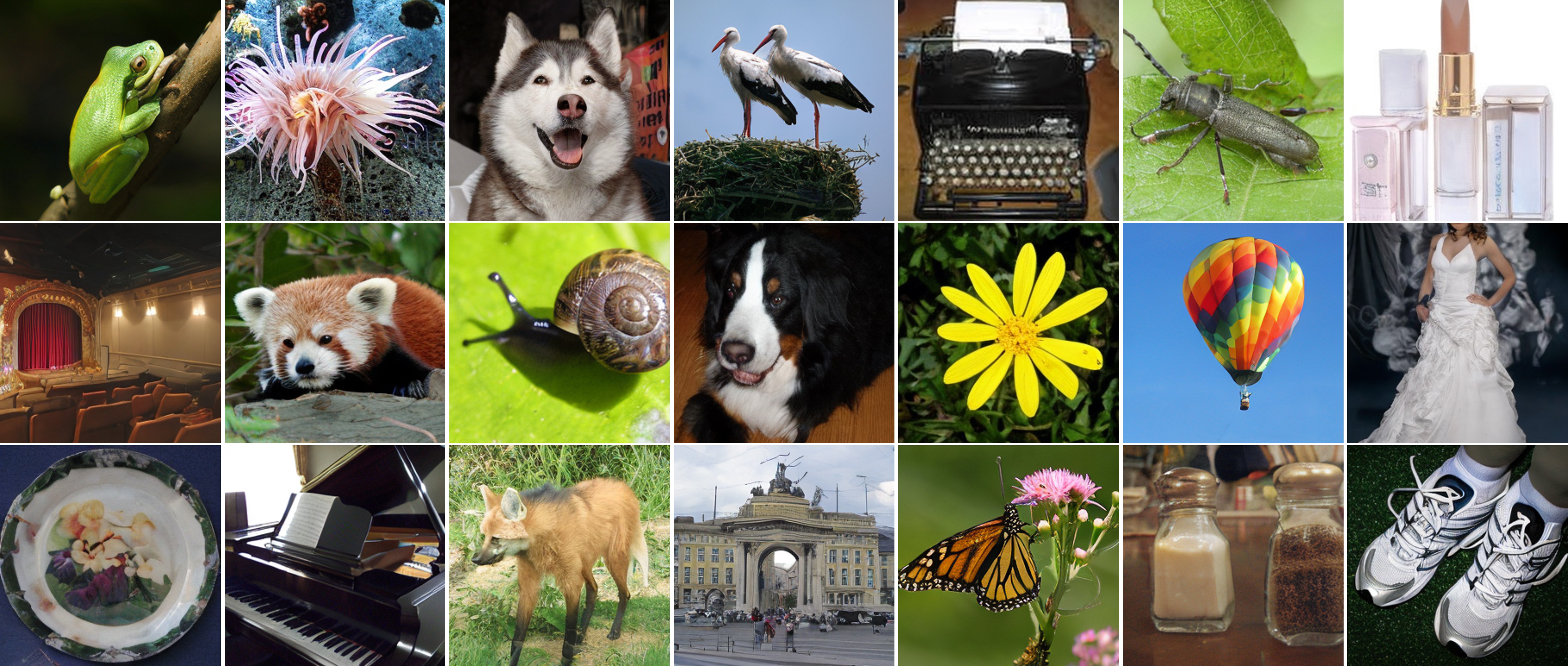}
    \vspace{-2em}
    \caption{Random generated $256\times 256$ samples from RaceDiT-XL/2-4in32. Classifier-free guidance scale is 4.}
    \label{fig:random_sample}
    \vspace{-0.5em}
\end{figure*}

We present additional results on the ImageNet $256\times256$. Our model is capable of generating high-quality images, as evidenced by the random samples shown in \cref{fig:random_sample} and the single-class generation results presented in \cref{sec:more_image}.
We further provide a comparison with leading approaches shown in \cref{tab:XL}, where \textit{Samples} is reported as \textit{training iterations} $\times$ \textit{batch size}. The classifier-free guidance scale is $1.5$ for evaluation metrics and $4.0$ for image generation results.

In our experiments, we train a vanilla DiT (marked with *) and a MoE model with Expert Race following the training protocol from the original DiT paper~\citep{dit}, except that we use a larger batch size {\bf 1024} to improve memory utilization and train for only {\bf 1.75M} steps. The learning rate remains unchanged at $1e^{-4}$. We also provide the training curves in~\cref{fig:teaser} (Right). Our MoE model, of similar amount of activated parameters, achieves better performance and faster convergence.

\begin{table*}[h]
    \vspace{-1em}
    \footnotesize
    \centering
    \caption{Comparison with other methods on ImageNet 256x256.}
    \vspace{-1em}
    \begin{tabular}{l|ccl|cccc}
    \hline
    \bf{Model Config.} & \bf{Total} & \bf{Activated} & \bf{Samples} & \bf{FID$\downarrow$} &  \bf{IS$\uparrow$} & \bf{Precision$\uparrow$} & \bf{Recall$\uparrow$} \\ \hline \hline
    ADM-G~\citep{cg} & 0.608B & 0.608B & 2.0M $\times$ 256  & 4.59  & 186.70 & 0.82 & 0.52 \\
    LDM-8-G~\citep{LDM} & 0.506B & 0.506B & 4.8M $\times$ 64 & 7.76  & 209.52 & 0.84 & 0.35 \\
    MDT~\citep{mdt} & 0.675B & 0.675B & 2.5M $\times$ 256 & 2.15  & 249.27 & 0.82 & 0.58 \\
    MDT~\citep{mdt} & 0.675B & 0.675B & 6.5M $\times$ 256 & 1.79  & 283.01 & 0.81 & 0.61 \\
    DiT-XL/2~\citep{dit} & 0.669B & 0.669B & 7.0M $\times$ 256 & 2.27  & 278.24 & 0.83 & 0.57 \\
    SiT-XL~\citep{sit} & 0.669B & 0.669B & 7.0M $\times$ 256 & 2.06  & 277.50 & 0.83 & 0.59 \\
    MaskDiT~\citep{maskdit} & 0.737B & 0.737B & 2.0M $\times$ 1024 & 2.50  & 256.27 & 0.83 & 0.56 \\ 
    DiT-MoE-XL/2~\citep{dit_moe} & 4.105B & 1.530B & 7.0M $\times$ 1024 & 1.72 &  315.73 & 0.83 & 0.64 \\
    \hline \hline
    DiT-XL/2* & 0.669B & 0.669B & 1.7M $\times$ 1024 & 3.02  & 261.49 & 0.81 & 0.51 \\
    RaceDiT-XL/2-4in32 & 2.788B & 0.707B & 1.7M $\times$ 1024 & 2.06  & 318.64 & 0.83 & 0.60 \\
    \hline \hline
    \end{tabular}
    \label{tab:XL}
\end{table*}
\vspace{-1em}

\section{Conclusion}
\vspace{-0.5em}

This paper proposes Expert Race, a novel Mixture-of-Experts (MoE) routing strategy that enables stable and efficient scaling of diffusion transformers. Compared to previous methods with fixed degrees of freedom in expert-token assignments, our strategy achieves higher routing flexibility by enabling top-k selection across the full routing space spanning batch, sequence, and expert dimensions. This expanded selection capability provides greater optimization freedom, significantly improving performance when scaling diffusion transformers. To address challenges from increased flexibility, we propose an EMA-based threshold adaptation mechanism that mitigates timestep-induced distribution shifts between training (randomized per-sample timesteps) and inference (uniform timesteps), ensuring generation consistency. Additionally, per-layer regularization improves training stability, while router similarity loss promotes diverse expert combinations and better load balancing, as shown on 256$\times$256 ImageNet generation tasks. As a general routing strategy, future work will extend Expert Race to broader diffusion-based visual tasks.

%% file: sections/appendix.tex
\section{Implemention of the Per-Layer Regularization}
\label{app:reg}
\begin{figure*}[h]
\centering
    \includegraphics[width=0.65\textwidth]{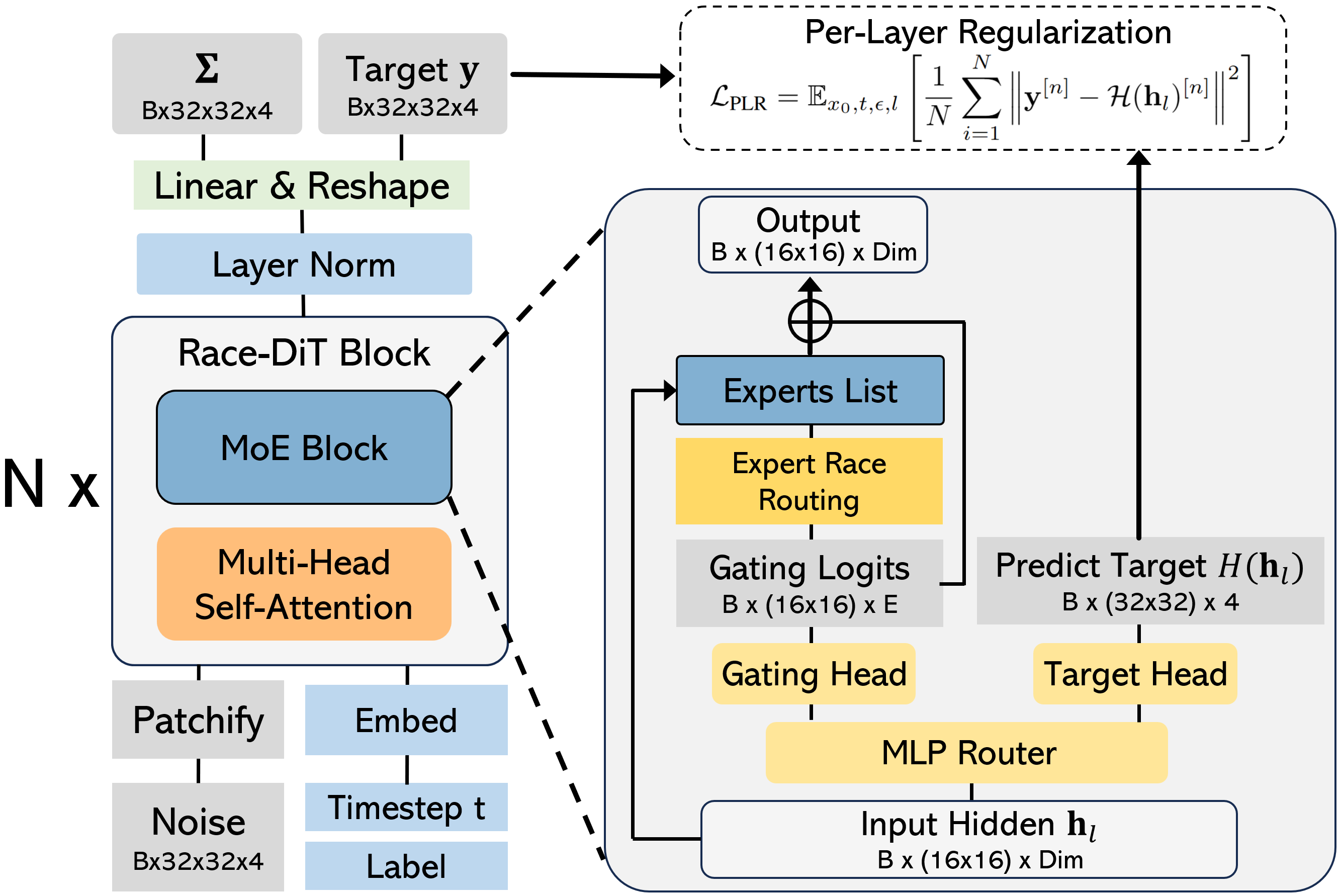}
    \caption{An illustration of the calculation for Per-Layer Regularization.}
    \label{fig:plr_illustration}
\end{figure*}

\cref{fig:plr_illustration} illustrates the Per-Layer Regularization mechanism. The input $\mathbf{h}_l$ at layer $l$ passes through a two-layer MLP router. The first layer applies linear transformation and GELU activation while preserving the original hidden dimension. The second layer branches into two heads: (1) a \textit{gating head} producing routing logits, and (2) a \textit{target head} predicting the output target $\mathbf{y}$ through linear projection. This dual-head design enables concurrent routing and target prediction. The L2 loss is computed per-layer between the target head's prediction $\mathcal{H}(\mathbf{h}_l)$ and ground truth $\mathbf{y}$. By aligning intermediate layer outputs with the final target, this regularization effectively enhances the contribution of shallow MoE layers to the final results in pre-norm architectures, significantly accelerating their optimization.

\section{Analysis of Router Similarity Loss}
\label{app:sim_loss}

In this section, we will demonstrate that the router similarity loss is an extended form of the balance loss, expanding from constraining individual experts to combinations of experts.

Load balance loss~\citep{deepseekmoe} are designed to encourage tokens to be evenly distributed across experts. For an MoE configuration with $E$ experts and a top-k value of $K$, the load balance loss for a batch with $T$ tokens is as follows:
\begin{equation}
\label{eq:blc}
\begin{aligned}
    \mathcal{L}_{blc} &= \sum^E_{i=1} f_i \cdot P_i 
    \\& = \sum^E_{i=1}\left(\frac{E} {KT}\sum_{t=1}^{T} \mathds{1}(i, t)\right) \left(\frac{1}{T} \sum_{t=1}^{T} s(i, t) \right)
\end{aligned}
\end{equation}
where $\mathds{1}(i, t)$ is the indicator function that equals $1$ when token $t$ is assigned to expert $i$ and $s(i,t)$ is the router logits of token $t$ and expert $i$.
Since softmax activation is applied over the expert dimension, $s(i,t)$ can be regarded as the probability that expert $i$ is selected given token $t$ as condition, marked as $p(i|t)$. 
Considering that each token is uniformly sampled from the dataset, $$\frac{1}{T} \sum_{t=1}^{T} s(i, t) = \sum_{t=1}^{T} p(i|t) \cdot p(t) = P_i$$ denotes to the marginal probability of choosing expert $i$.
And $$f_i = \frac{E}{KT}\sum_{t=1}^{T} \mathds{1}(i, t) = \frac{\sum_{t=1}^{T} \mathds{1}(i, t)}{\frac{1}{E} \sum_{i=1}^{E} \sum_{t=1}^{T} \mathds{1}(i, t)}$$ 
is the ratio of the actual number of tokens assigned to expert $i$ to the expected number of tokens assigned to each expert.

As for router similarity loss, each element of matrix $P'$ in \cref{eq:matrix} can be formulated as below:
\begin{equation}
    \begin{aligned}
    \frac{1}{T} P'_{i,j} &= \frac{1}{T} \sum_{t=1}^{T} s(i, t) \cdot s(j,t) \\
    &= \sum_{t=1}^{T} p(i|t) \cdot p(j|t) \cdot p(t) \\
    &= p(i,j)
\end{aligned}
\end{equation}
The matrix $P'$ actually represents the probability of the pair of experts 
$i$ and $j$ being selected.
Similarly, each element in $M'$ of \cref{eq:matrix} is the ratio of the actual number of tokens assigned to the expert pair $(i, j)$ to the expected number of tokens assigned to each pairwise combination and thus $W(i,j)$ of \cref{eq:sim_loss} has a similar form to $f_i$ in balance loss.

Noted that the above equations model the probability of sampling with replacement for experts. However, in practical implementation, sampling is performed without replacement, and the case that one expert is selected more than once does not exist. Therefore, in \cref{eq:sim_loss_weight}, we normalize the diagonal and off-diagonal parts of the matrix separately. The diagonal part degenerates into an estimation for individual experts as follows:
\begin{equation}
    \begin{aligned}
        \mathcal{L}_{sim\_diag} &= \frac{1}{T} \sum_{i=1}^{E} W(i,i) \cdot P'_{i,i} \\
        &= \frac{E}{T} \cdot \sum_{i=1}^{E} \frac{M'_{i,i}}{\sum_{j}^{E} M'_{j,j}} \cdot \sum_{t=1}^{T} s(i,t)^2 \\
        &= \frac{E}{T} \cdot \sum_{i=1}^{E} \frac{\sum_{t=1}^{T} \mathds{1}(i, t)^2}{\sum_{j=1}^{E} \sum_{t=1}^{T} \mathds{1}(j, t)^2} \cdot \sum_{t=1}^{T} s(i,t)^2 \\
        &= \frac{E}{T} \cdot \sum_{i=1}^{E} \frac{\sum_{t=1}^{T} \mathds{1}(i, t)}{\sum_{j=1}^{E} \sum_{t=1}^{T}\mathds{1}(j, t)} \cdot \sum_{t=1}^{T} s(i,t)^2 \\
        &= \frac{E}{T} \cdot \sum_{i=1}^{E} \frac{\sum_{i=1}^{T} \mathds{1}(i, t)}{KT} \cdot \sum_{t=1}^{T} s(i,t)^2 \\
        &= \sum_{i=1}^{E} \left( \frac{E}{KT} \sum_{t=1}^{T} \mathds{1}(i, t) \right) \left( \frac{1}{T} \sum_{t=1}^{T} s(i,t)^2 \right) \\
    \end{aligned}
\end{equation}

Therefore, the diagonal part of router similarity loss represents a geometric mean version of the balance loss as described above.
The $W(i,i)$ in the router similarity loss is identical to $f_i$ in the balance loss. Similarly, for the off-diagonal weights $W(i,j)$, we estimate the expected number of times that each expert pair is selected by dividing the sum of all elements, $\sum_{i \neq j} W(i,j)$, by the number of terms, $E^2 - E$.
The complete router similarity loss can be viewed as an augmentation of the traditional balance loss, particularly by incorporating the pairwise interaction between experts, to help the model make more effective use of different combinations of experts.
It is worth noting that we normalize both the weights and the probabilities, making the calculation of the loss function independent of the number of experts, top-k, and token sequence length. Considering a random assignment case, assume that the scores of the entire score matrix $S$ are the same constant. In this case, $P_{i,j} = \frac{T}{E^2}$, substituting this into the \cref{eq:sim_loss} yields $\mathcal{L}_{sim}=1$.

\section{Combination Usage}
\label{app:combine}
\begin{figure*}[h]
\centering
    \includegraphics[width=0.8\textwidth]{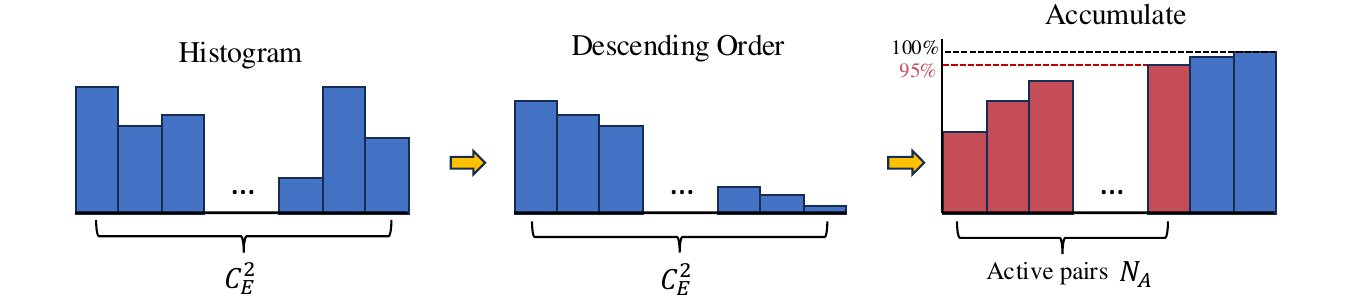}
    \caption{Compute process of Combination Usage. }
    \label{fig:combine_process}
\end{figure*}
To estimate the actual number of pairwise expert combinations activated and involved in computation, we use the metric called Combination Usage whose process is shown in the \cref{fig:combine_process}.
From the perspective of pairwise combinations, for $E$ experts, there are a total of ${E \choose 2}$ pairwise combinations. 
First, we count the number of times each expert pair is selected across all tokens, resulting in a histogram. Then, we sort the counts in descending order and normalize them to obtain a sorted normalized histogram.
Finally, we compute the cumulative sum and count the number of bins whose cumulative sum is less than 95\%.
Finally, we compute the cumulative sum and calculate the proportion of bins whose cumulative sum is less than 95\% relative to the total number of bins, \textit{i.e.}, $N_A / C_E^2$. This proportion is referred to as the Combination Usage.
In other words, those expert pairs with their amount less than 5\% of the total count are considered inactive.
We set k=4 and conducted experiments with different numbers of experts, as shown in the \cref{fig:combination}. The experiment results indicate that our method effectively enhances the utilization of a larger number of combinations.

\section{Additional Comparisons with DiT-MoE}
\vspace{-1em}
\begin{table*}[h]
    \small
    \centering
    \caption{Comparison to DiT-MoE}
    \begin{tabular}{l|ccc|ccc}
    \hline
    \bf{Model Config.}           & \bf{Total Param.} & \bf{Activated Param.} & \bf{Iters.} & \bf{FID$\downarrow$}      & \bf{CMMD$\downarrow$}     & \bf{CLIP$\uparrow$}     \\ \hline\hline
    DiT-MoE-B/2-8E2A~\citep{dit_moe}       & 0.795B            & 0.286B         & 500K       &  9.06                     & .5049                     & 22.87         \\
    RaceDiT-B/2-2in8           & 0.297B            & 0.135B               & 500K &  8.02                     & .4587                     & 23.09         \\
     \hline\hline
    \end{tabular}
    \label{tab:ditmoe_comp}
\end{table*}
\vspace{-1em}
\begin{figure*}[h!]
\centering
    \includegraphics[width=0.4\textwidth]{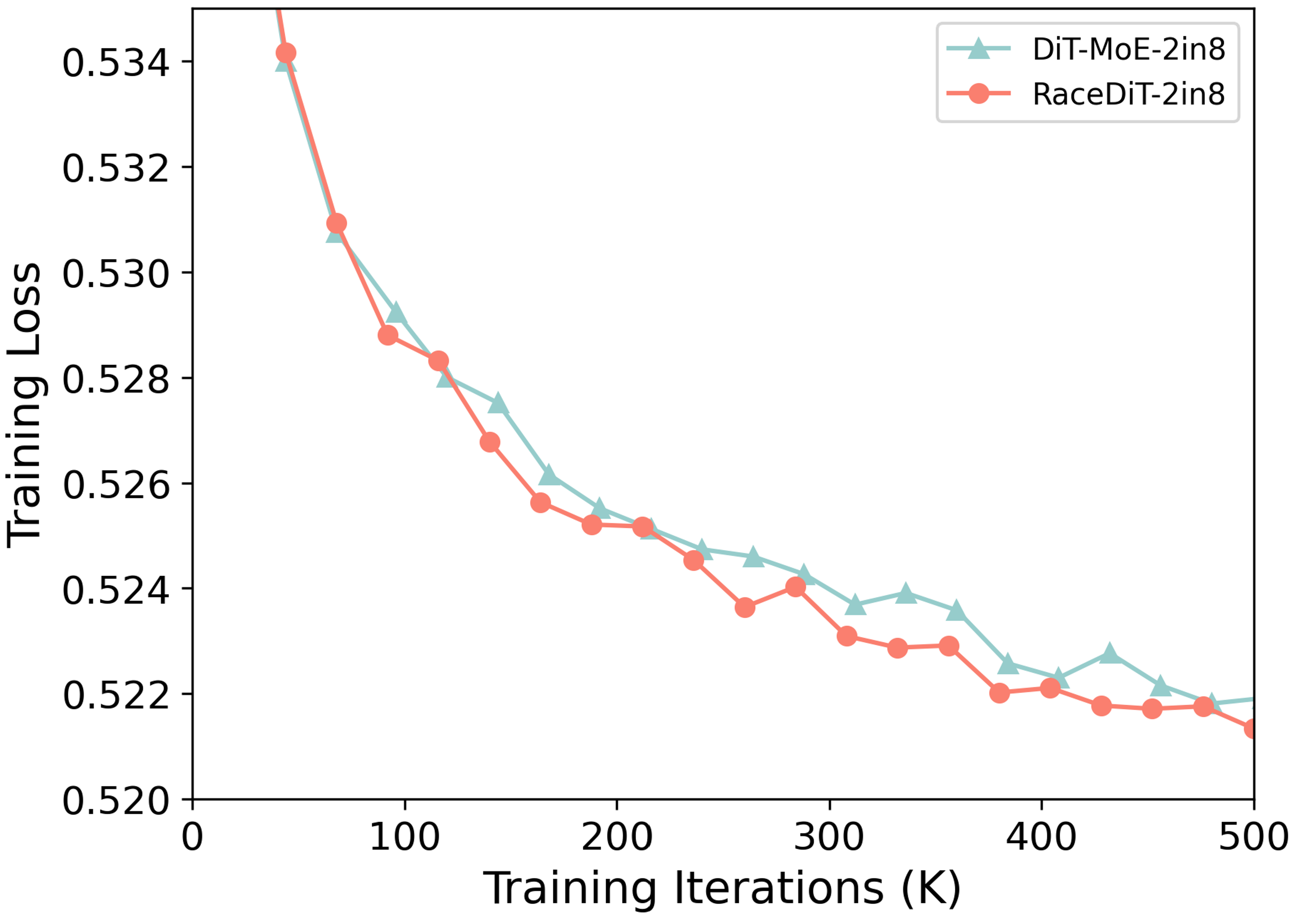}
    \vspace{-1em}
    \caption{Training curve comparisons between DiT-MoE~\citep{dit_moe} and our model.}
    \label{fig:train_curve_ditmoe}
\end{figure*}

We provide an additional comparison with DiT-MoE~\citep{dit_moe}, our most-related work using its official open-source code. We conduct experiments under the DiT-MoE-B/2-8E2A setting and compared it with our RaceDiT-B/2-2in8 model using the same training configuration. Both models were trained for 500K iterations. 

There are several differences between the models: DiT-MoE uses GLU, while our method employs a standard MLP. Additionally, DiT-MoE includes two extra shared experts, which our model does not. As a result, our model has fewer total and activated parameters. Despite these differences, as demonstrated in \cref{tab:ditmoe_comp} and \cref{fig:train_curve_ditmoe}, our method achieves better training loss and evaluation metrics, even with a smaller number of activated parameters.

\newpage
\section{Additional Image Generation Results} \label{sec:more_image}
\begin{multicols}{2}
    \begin{figure}[H]
        \centering
        \includegraphics[width=.95\linewidth]{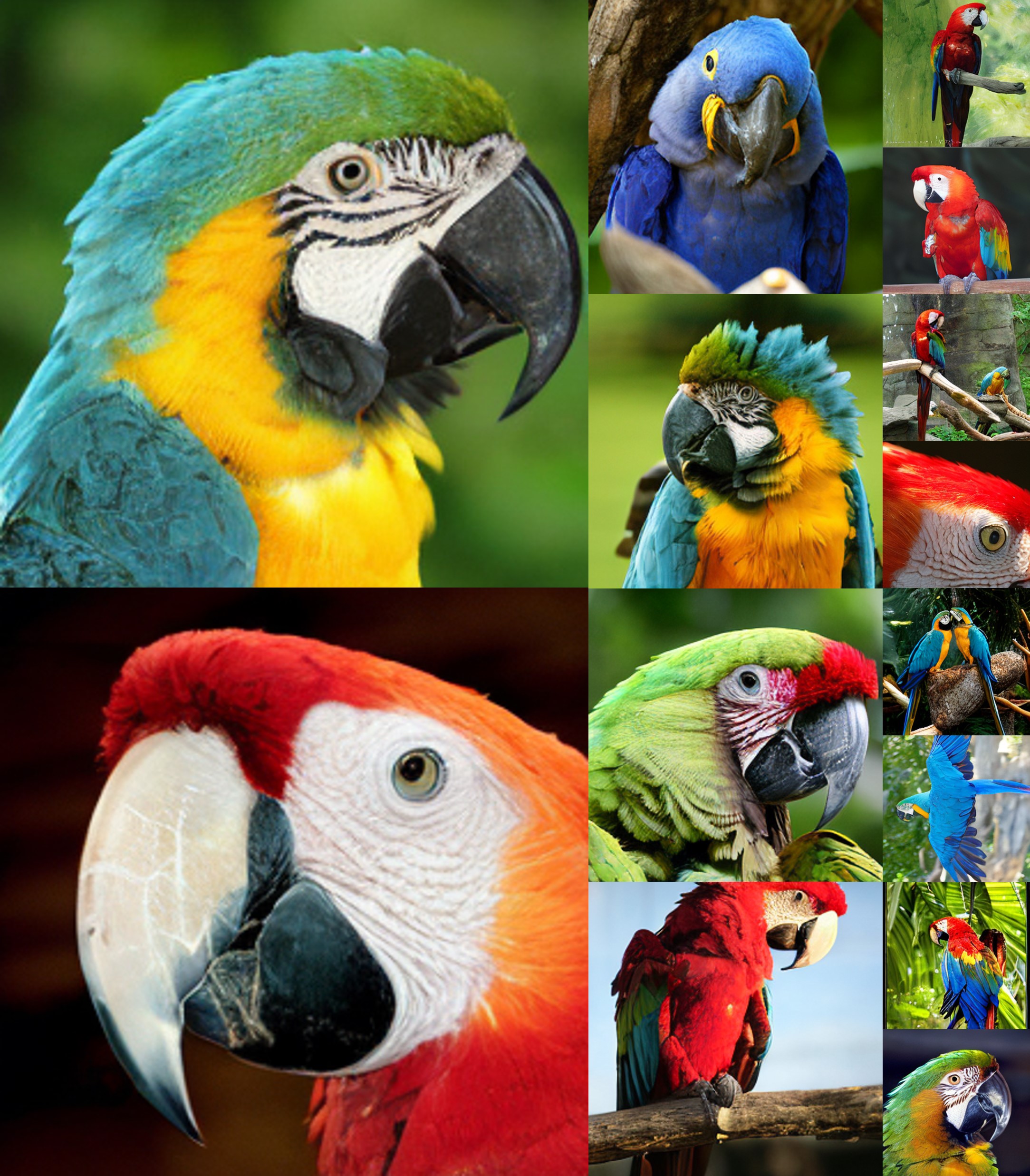}
        \caption{Samples from RaceDiT-XL/2-4in32 ($256\times256$).\\
        Classifier-free guidance: 4.0\\
        Label: Macaw (88)
        }
    \end{figure}
    \begin{figure}[H]
        \centering
        \includegraphics[width=.95\linewidth]{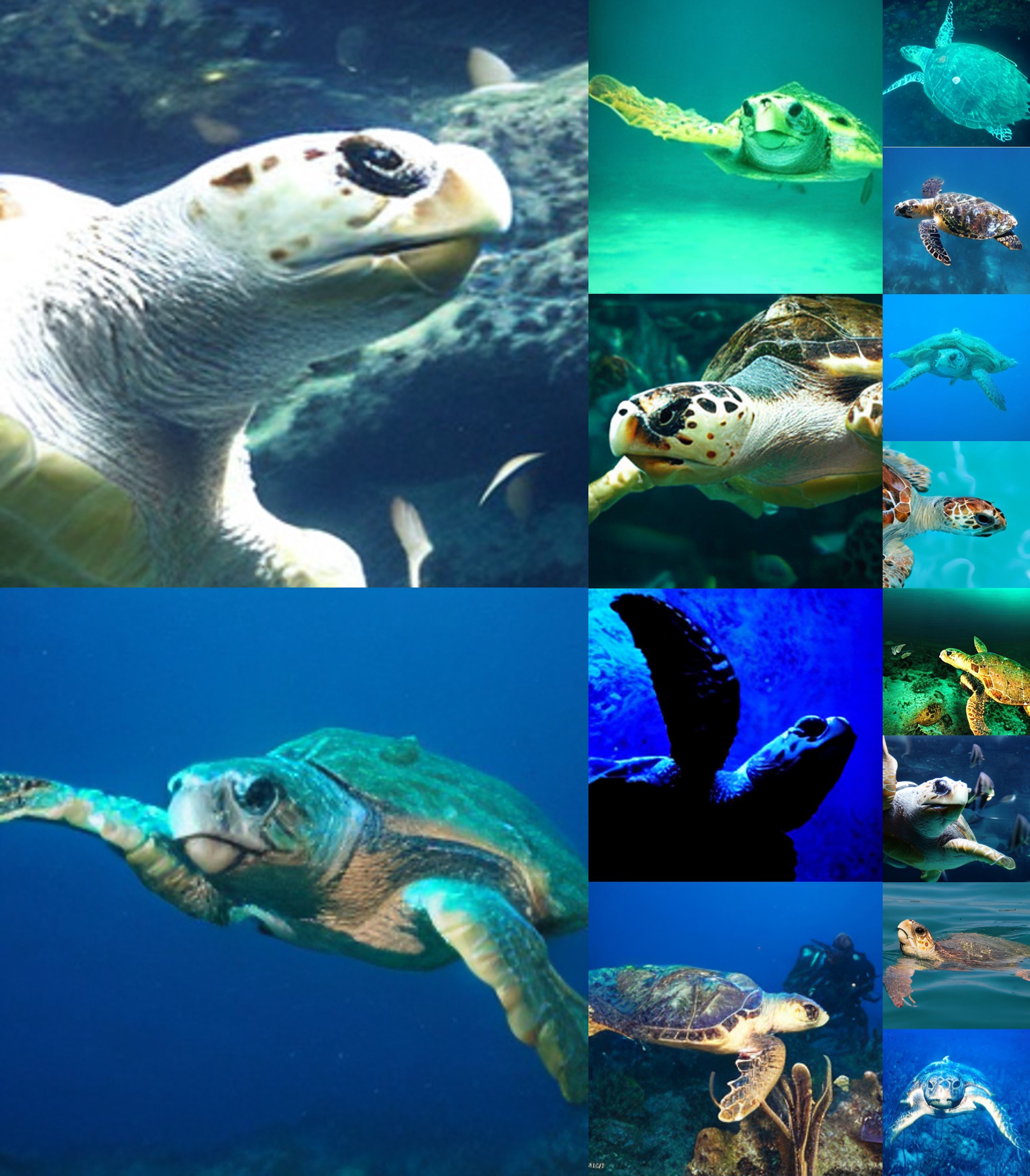}
        \caption{Samples from RaceDiT-XL/2-4in32 ($256\times256$).\\
        Classifier-free guidance: 4.0\\
        Label: loggerhead turtle (33)
        }
    \end{figure}
\end{multicols}

\begin{multicols}{2}
    \begin{figure}[H]
        \centering
        \includegraphics[width=.95\linewidth]{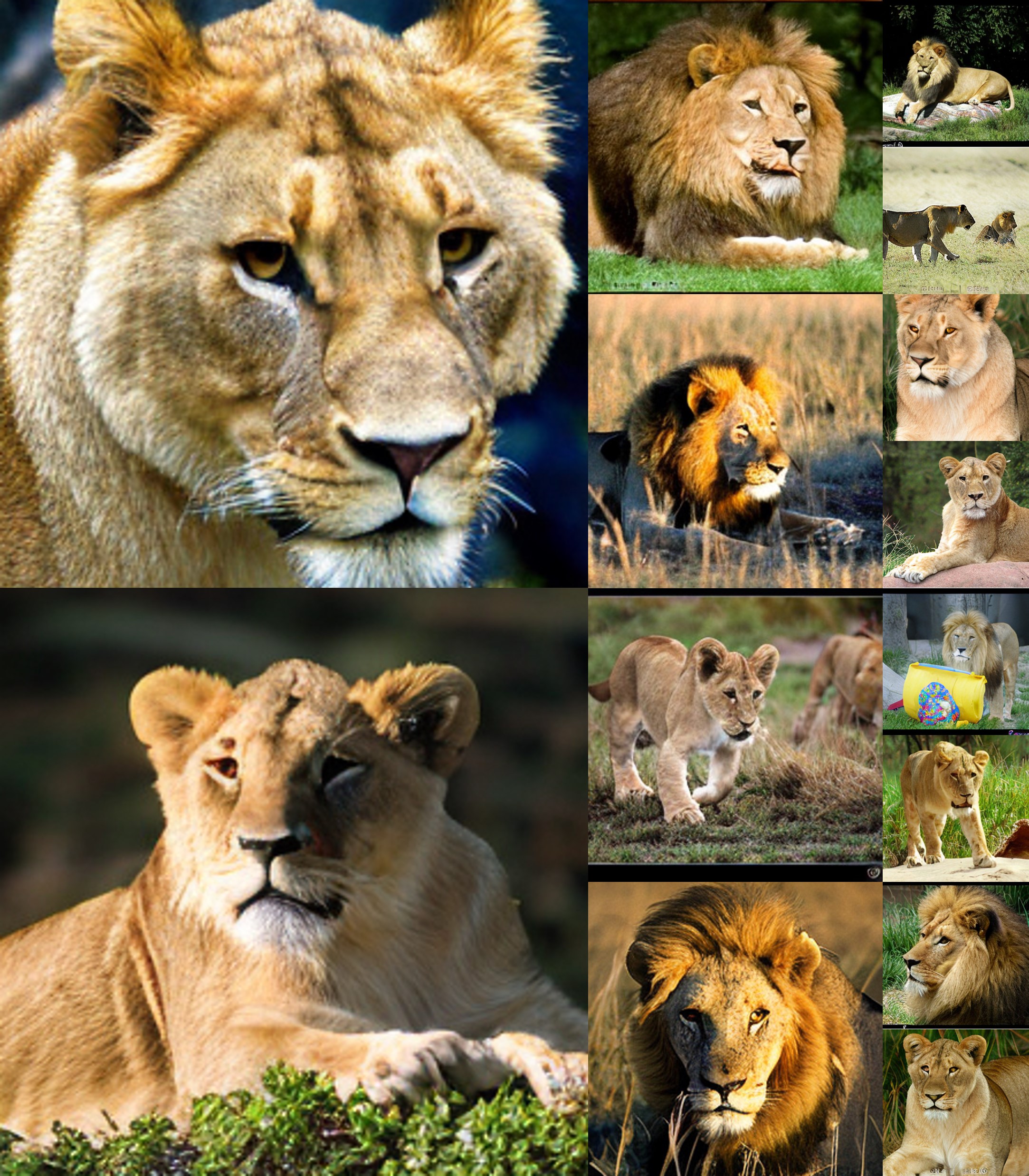}
        \caption{Samples from RaceDiT-XL/2-4in32 ($256\times256$).\\
        Classifier-free guidance: 4.0\\
        Label: lion (291)
        }
    \end{figure}
    \begin{figure}[H]
        \centering
        \includegraphics[width=.95\linewidth]{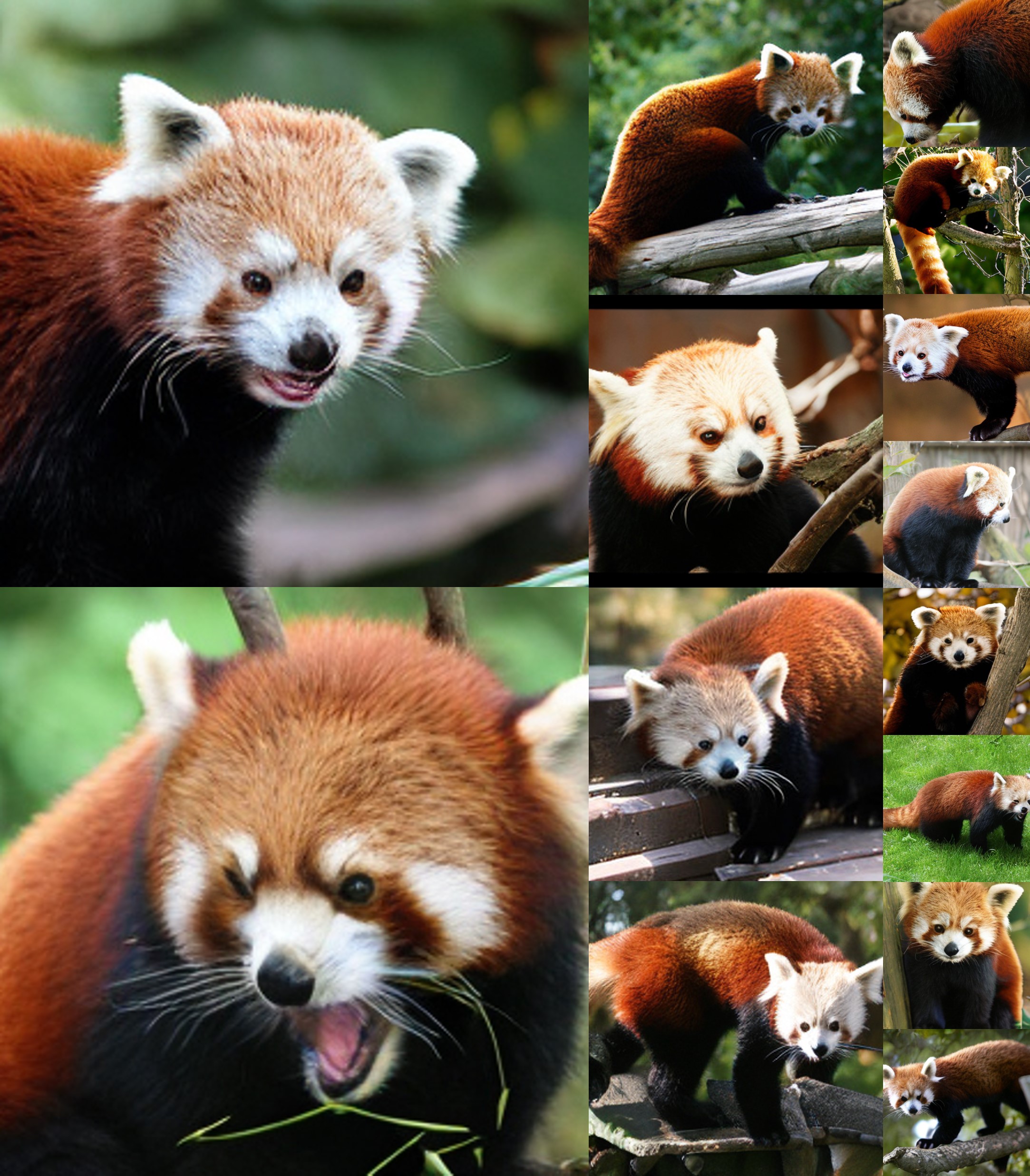}
        \caption{Samples from RaceDiT-XL/2-4in32 ($256\times256$).\\
        Classifier-free guidance: 4.0\\
        Label: lesser panda (387)
        }
    \end{figure}
\end{multicols}

\begin{multicols}{2}
    \begin{figure}[H]
        \centering
        \includegraphics[width=.95\linewidth]{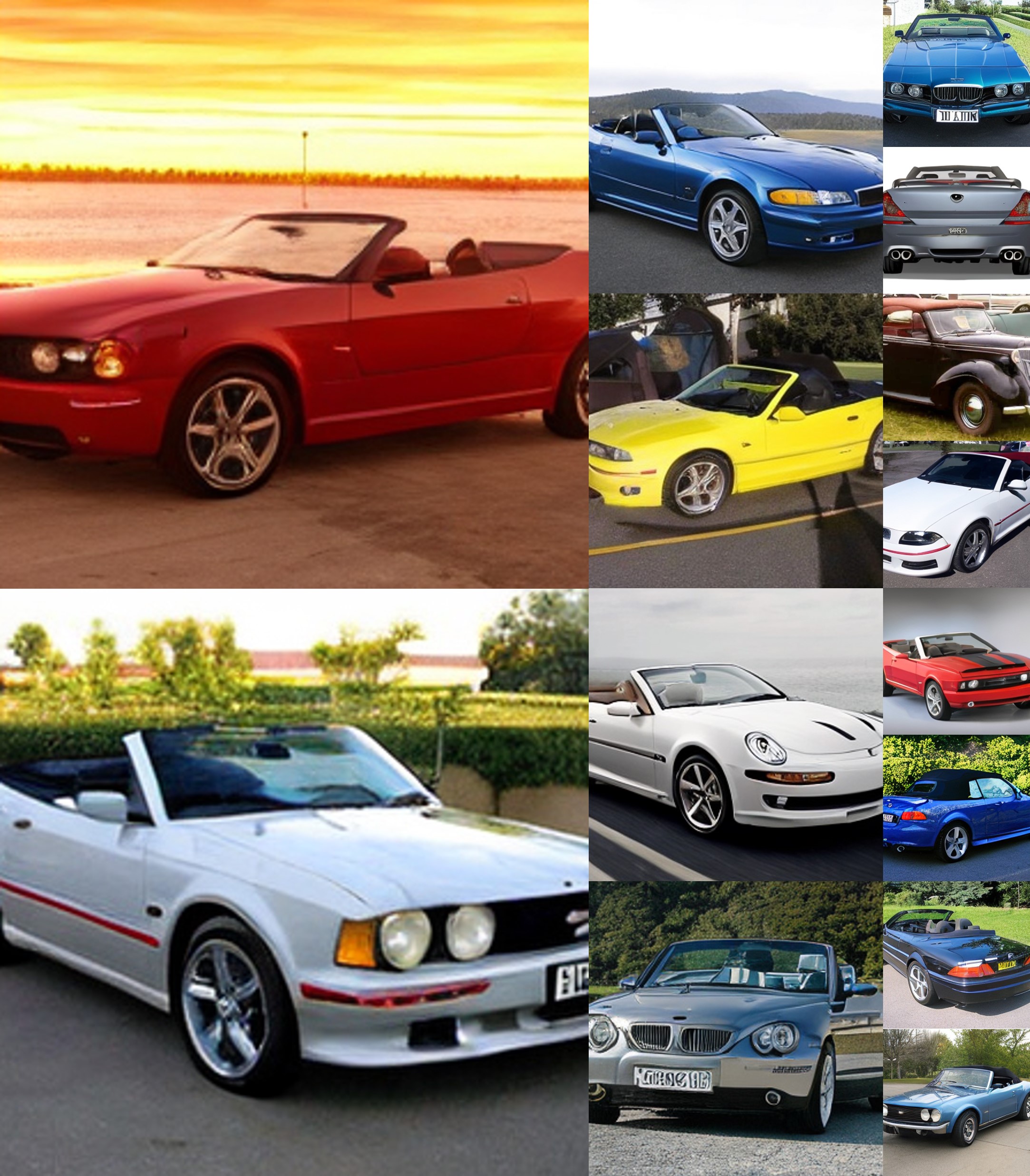}
        \caption{Samples from RaceDiT-XL/2-4in32 ($256\times256$).\\
        Classifier-free guidance: 4.0\\
        Label: convertible (511)
        }
    \end{figure}
    \begin{figure}[H]
        \centering
        \includegraphics[width=.95\linewidth]{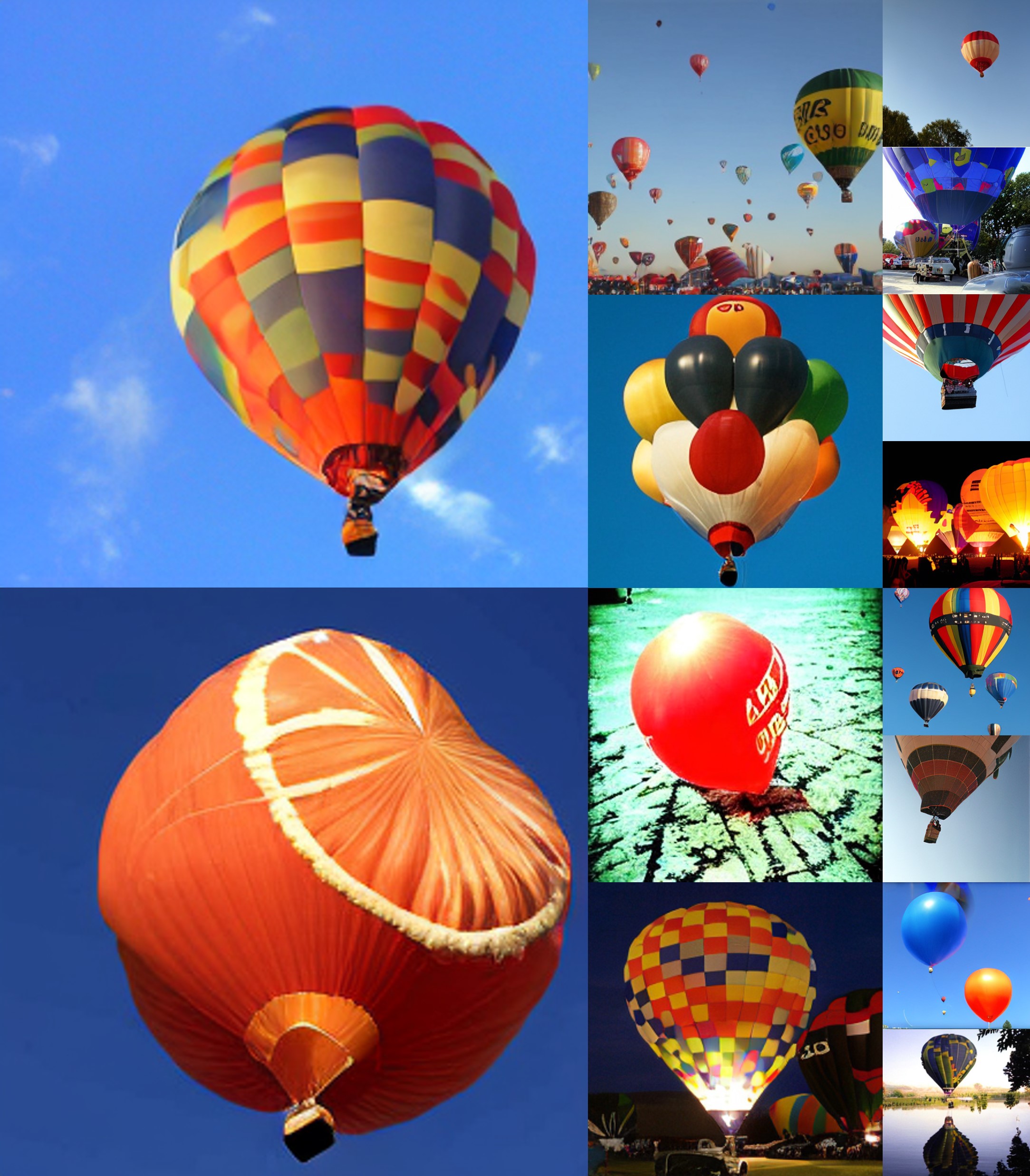}
        \caption{Samples from RaceDiT-XL/2-4in32($256\times256$).\\
        Classifier-free guidance: 4.0\\
        Label: balloon (417)
        }
    \end{figure}
\end{multicols}

\begin{multicols}{2}
    \begin{figure}[H]
        \centering
        \includegraphics[width=.95\linewidth]{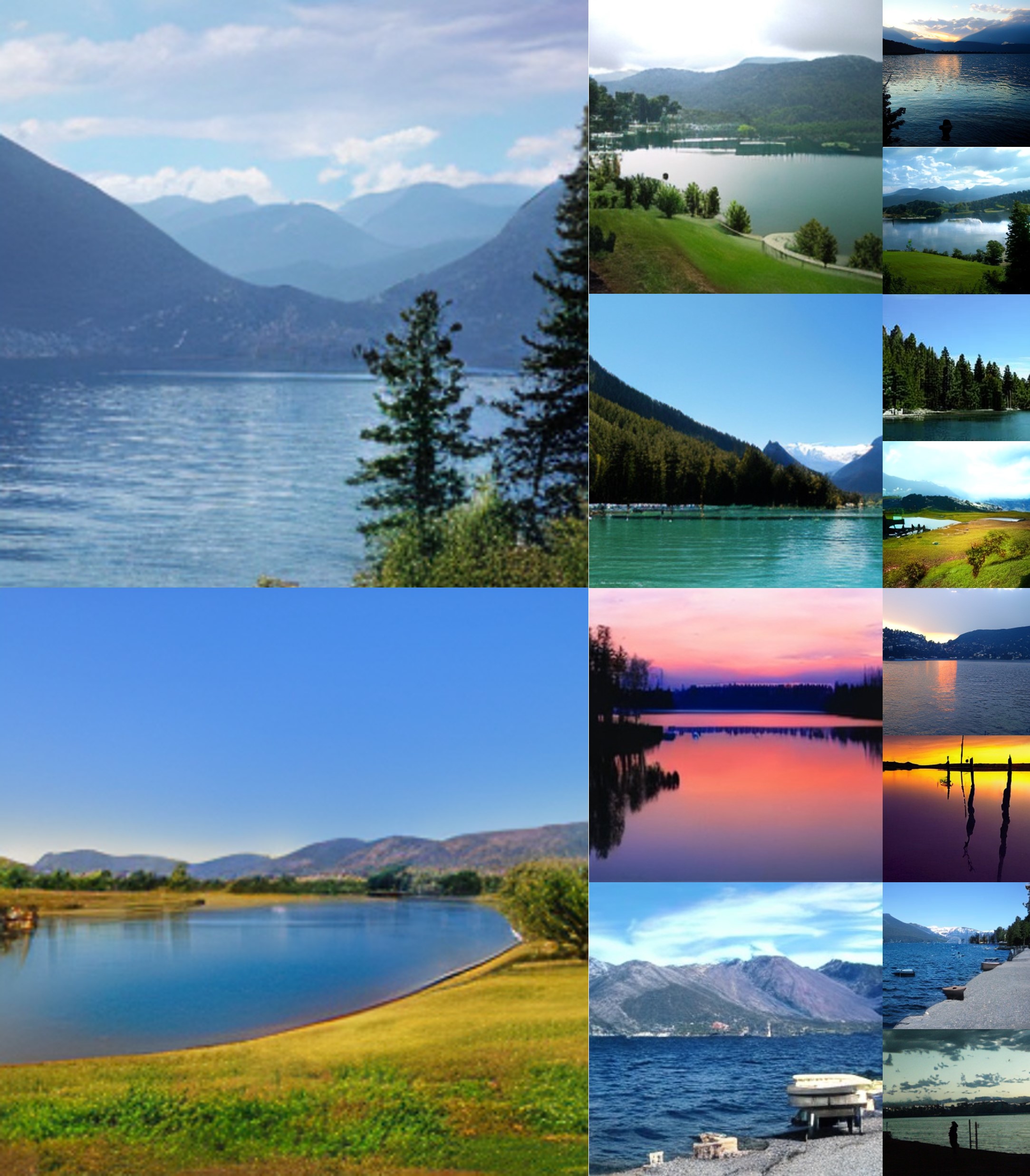}
        \caption{Samples from RaceDiT-XL/2-4in32 ($256\times256$).\\
        Classifier-free guidance: 4.0\\
        Label: lakeside (975)
        }
    \end{figure}
    \begin{figure}[H]
        \centering
        \includegraphics[width=.95\linewidth]{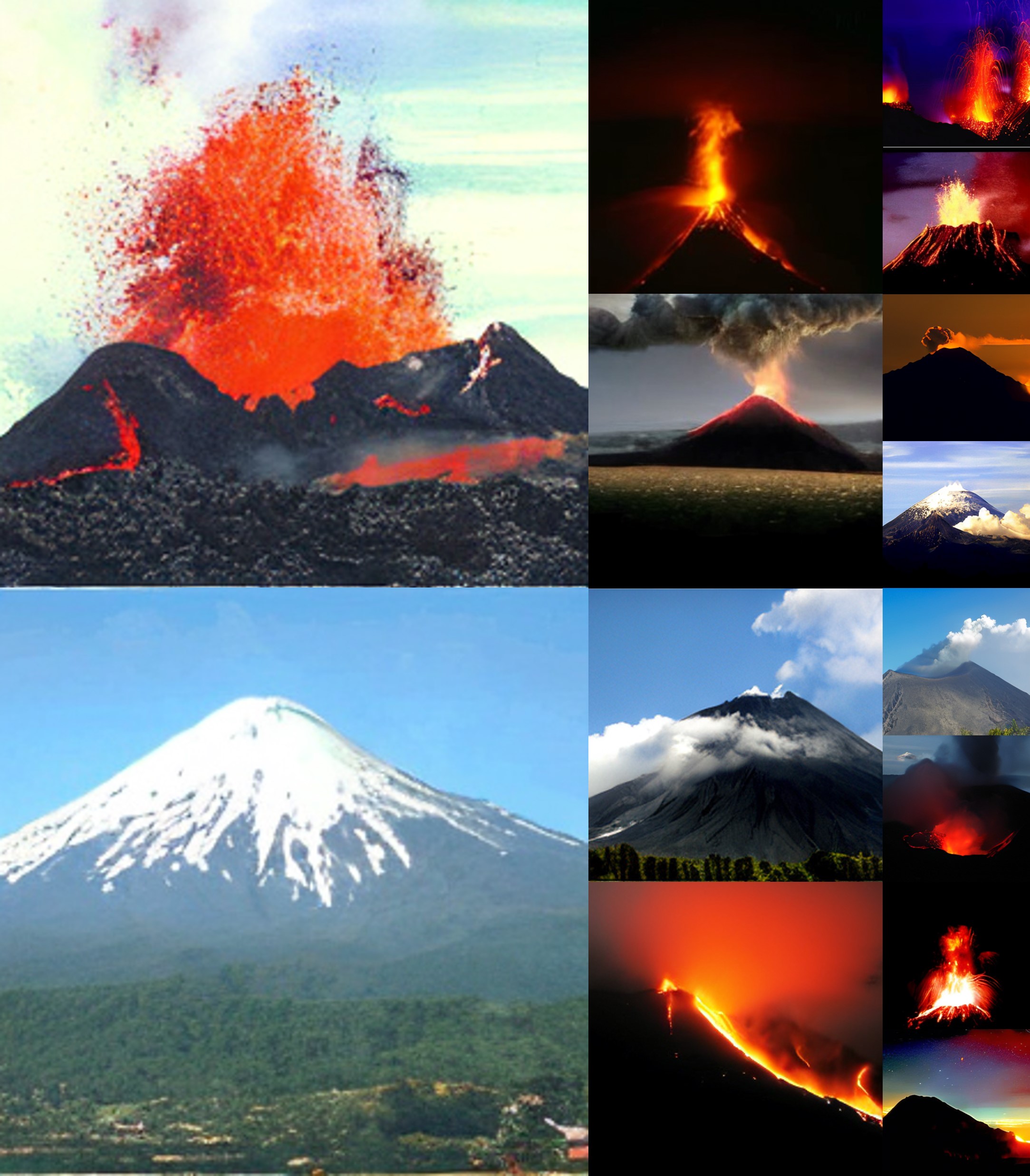}
        \caption{Samples from RaceDiT-XL/2-4in32 ($256\times256$).\\
        Classifier-free guidance: 4.0\\
        Label: volcano (980)
        }
    \end{figure}
\end{multicols}